\documentclass[10pt,twocolumn,letterpaper]{article}

\usepackage{cvpr}
\usepackage{times}
\usepackage{epsfig}
\usepackage{graphicx}
\usepackage{amsmath}
\usepackage{amssymb}
\usepackage{array}

\usepackage{authblk} 

\usepackage{color}

\usepackage[svgnames]{xcolor}
\usepackage{soul} 

\newcommand{\hlc}[2][yellow]{ {\sethlcolor{#1} \hl{#2}} }
\usepackage{enumitem}

\usepackage{fullpage}
\usepackage{times}
\usepackage{fancyhdr,graphicx,amsmath,amssymb}
\usepackage[linesnumbered,ruled,vlined]{algorithm2e}
\include{pythonlisting}

\SetCommentSty{mycommfont}

\SetKwInput{KwInput}{Input}                
\SetKwInput{KwOutput}{Output}              

\usepackage{caption, subcaption} 

\usepackage{hhline} 

\usepackage{comment}

\usepackage{array}
\newcolumntype{L}[1]{>{\raggedright\let\newline\\\arraybackslash\hspace{0pt}}m{#1}}
\newcolumntype{C}[1]{>{\centering\let\newline\\\arraybackslash\hspace{0pt}}m{#1}}
\newcolumntype{R}[1]{>{\raggedleft\let\newline\\\arraybackslash\hspace{0pt}}m{#1}}

\usepackage[labelsep=period]{caption}

\usepackage[pagebackref=true,breaklinks=true,letterpaper=true,colorlinks,bookmarks=false]{hyperref}

\cvprfinalcopy 

\begin{document}

\title{TGGLines: A Robust Topological Graph Guided \\ Line Segment Detector for Low Quality Binary Images}

\author[1]{Ming Gong}
\author[2,3*]{Liping Yang}
\author[4]{Catherine Potts}
\author[1]{Vijayan K. Asari} 
\author[2]{\\Diane Oyen}
\author[2]{Brendt Wohlberg}
\affil[1]{University of Dayton, Dayton, OH, USA}
\affil[2]{Los Alamos National Laboratory, Los Alamos, NM, USA}
\affil[3]{University of New Mexico, Albuquerque, NM, USA}
\affil[4]{Montana State University, Bozeman, MT, USA}
\affil[*]{\textbf{Corresponding author: }\textit{Liping Yang}, lipingyang@unm.edu}





\maketitle

\begin{abstract}
   Line segment detection is an essential task in computer vision and image analysis, as it is the critical foundation for advanced tasks such as shape modeling and road lane line detection for autonomous driving. We present a robust topological graph guided approach for line segment detection in low quality binary images (hence, we call it TGGLines). Due to the graph-guided approach, TGGLines not only detects line segments, but also organizes the segments with a line segment connectivity graph, which means the topological relationships (e.g., intersection, an isolated line segment) of the detected line segments are captured and stored; whereas other line detectors only retain a collection of loose line segments. Our empirical results show that the TGGLines detector visually and quantitatively outperforms state-of-the-art line segment detection methods. In addition, our TGGLines approach has the following two competitive advantages: (1) our method only requires one parameter and it is adaptive, whereas almost all other line segment detection methods require multiple (non-adaptive) parameters, and (2) the line segments detected by TGGLines are organized by a line segment connectivity graph.


\end{abstract}


\section{Introduction}

\begin{figure*}[!htb]
\begin{center}
   \begin{subfigure}{0.3\textwidth}
      \includegraphics[width=\textwidth]{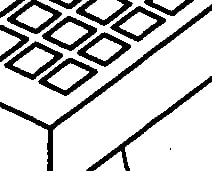}
      \caption{Input diagram image}
    \end{subfigure}
    \hfill
    \begin{subfigure}{0.3\textwidth}
      \includegraphics[width=\textwidth]{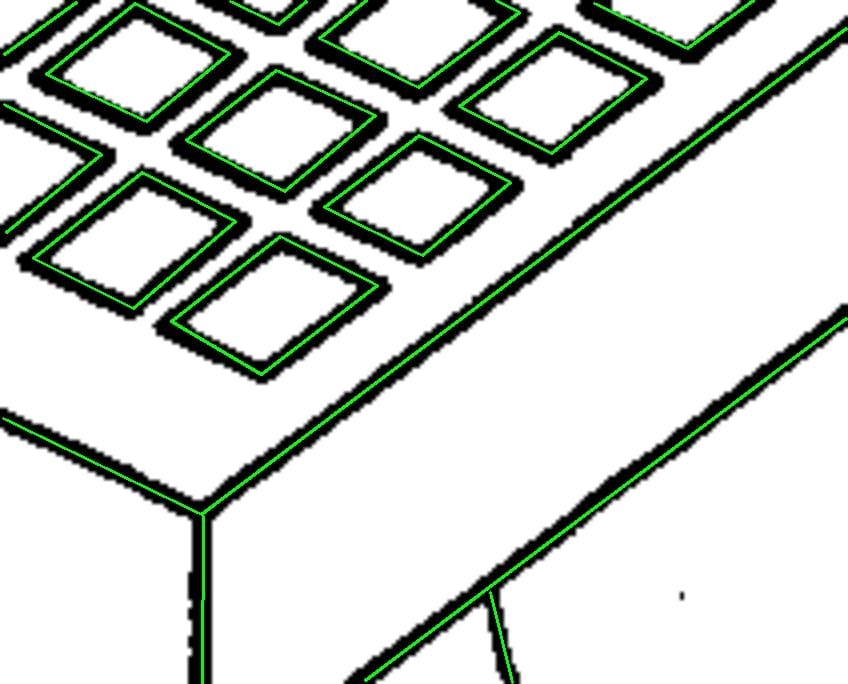}
      \caption{What humans see}
    \end{subfigure}
    \hfill
    \begin{subfigure}{0.32\textwidth}
      \includegraphics[width=\textwidth]{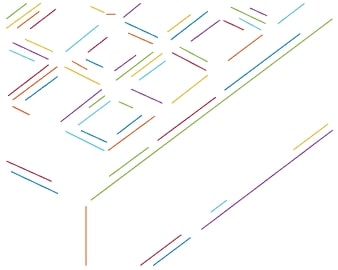}
      \caption{What machines see}
    \end{subfigure}
\end{center}
   \caption{An example of how challenging it is for machines to \textit{see} line segments present in diagram images (e.g., patent images), especially those with low quality introduced by the scanning process. (a) Input image is a binary pixel-raster image. (b) Line segments (perceived) and annotated in green by humans. (c) Line segments machines see in the input image (using state-of-art line segment detection method \emph{linelet} \cite{cho2017novel}).} 
\label{fig:lineseg_challenge_example}
\end{figure*}

Line segment detection has been studied for decades in computer vision and plays a fundamental and important role for advanced vision problems, such as indoor mapping \cite{an2012line},  vanishing point estimation \cite{lezama2014finding,xu2013minimum}, and road lane line detection for autonomous driving \cite{beyeler2014vision}.
Many line segment detection methods are developed for natural images \cite{ballard1981generalizing,matas2000robust,von2010lsd,akinlar2011edlines,cho2017novel}. However, those methods do not work well for technical diagram images, such as patent images -- see Figure \ref{fig:lineseg_challenge_example} for an example. Clearly the line segment detection challenge remains in technical images, especially in low quality images affected by the local zigzag ``noise'' introduced by the scanning process.

We present a robust topological graph guided approach for line segment detection in low quality binary (diagram) images. Specifically, our approach combines the power of topological graphs and skeletons to generate a skeleton graph representation to tackle the line segment detection challenges. A skeleton is a central line (1 pixel wide) representation of an object present in an image obtained via thinning \cite{lam1992thinning,Yang_2019_CVPR_Workshops}. The skeleton emphasizes topological and geometrical properties of shapes \cite{Yang_2019_CVPR_Workshops}. In our approach, the skeleton serves as the essential bridge from pixel image representation to topological graph representation.

We compare our TGGLines approach with four mainstream and state-of-the-art line segment detection methods. The empirical
results demonstrate that our TGGLines approach visually and quantitatively outperforms other methods. In addition, our method has two advantages. (1) While the parameters for most other methods are not adaptive, our method is robust as it only requires one parameter, and this parameter is \textit{adaptive}. (2) As TGGLines detects line segments it organizes the segments into a topological graph. We call this graph the \textit{line segment connectivity graph} (LSCG); which stores the topological relations (e.g., turning and junction) among detected line segments. The LSCG can be used by advanced computer vision tasks, such as shape analysis and line segment-based image retrieval.

Line detection performance is often simply evaluated by
visualization, which is a non-quantitative evaluation. In this paper, we evaluate our results both qualitatively (Section \ref{sec:evaluation_qualitative}) and quantitatively (Section \ref{sec:evaluation_quantitative}). One of the most difficult problems for quantitative evaluation is annotating line segments in an image accurately because sometimes even humans will annotate lines differently due to the zigzag ``noise" introduced by the scanning process (see the annotation examples provided in Figure \ref{fig:lineseg_annotation_examples}). Typically for a 
scientific diagram image, there can be several hundreds of lines that must be annotated (see images \#6 and \#10 in Table \ref{tab:result_images_part2} for examples). We provide a simple interface for line segment annotation as well as a quantifiable metric for line detection performance as measured with respect to inherently inexact annotations.

Here, we provide a road map to the rest of the paper. Section~\ref{sec:related_work} covers related work, including existing line segment detection methods, and the topological-based image representations that our method is built on. The core of our paper is Section~\ref{sec:approach} focusing on our TGGLines approach and Section~\ref{sec:tgg_algs} focusing on algorithms. 
In Section~\ref{sec:experiments_evaluation}, we present our experiments with qualitative and quantitative evaluations. The paper concludes in Section~\ref{sec:conclusion} with a mention of potential applications.

For readability, we provide a list of abbreviations in Appendix~\ref{app:abbr}. Background on graph theory and computational geometry, is provided in Appendix~\ref{app:def_graph_terms}. These appendices are provided in the supplementary materials.

\section{Related work}\label{sec:related_work}


Existing line segment detection methods are either edge-based or local gradient-based, whereas our TGGLines method does not rely on edge detection or local gradients. Edge-based line detectors include the \textit{Hough transform (HT)} \cite{ballard1981generalizing} which often uses the Canny edge detector \cite{canny1986computational} as pre-processing. The main drawback of HT is that it is computationally expensive and it only outputs the parameters of a line equation, not the endpoints of line segments. The \textit{progressive probabilistic Hough transform (PPHT)} \cite{matas2000robust} is an optimization of the standard HT; it does not take all the points into consideration, instead taking only a random subset of points and that is sufficient for line detection. Also, PPHT is able to detect the two endpoints of each line, so it can be used to detect line segments present in images.



Local gradient-based line detectors are successful on natural images, but not diagrams (See Section \ref{sec:experiments_results}). The \textit{line segment detector (LSD)} \cite{von2010lsd} is a local gradient-based method that has been tested on a wide set of natural images.
\textit{EDLines} \cite{akinlar2011edlines} speeds up LSD but according to our experiments and evaluation (Section \ref{sec:experiments_results}), EDLines' performance is much worse than LSD on the diagram image dataset. 
The \textit{linelet} \cite{cho2017novel} method represents intrinsic properties of line segments in images using an undirected graph which performs well on an urban scene dataset, but does not work well for diagrams.
 A recent review about line segment detection methods can be found in \cite{guerreiro2012connectivity,rahmdel2015review}.
 


Our TGGLines method builds on topological-based methods. TGGLines uses the skeleton graph image representation proposed in \cite{mlg2019_7}. The topological graph is generated automatically from an image skeleton, which can capture the topological and geometrical properties of shapes of objects present in an image \cite{lakshmi2009survey}. We use the well-known and robust Zhang-Suen thinning algorithm \cite{zhang1984fast} to extract skeletons from images and simplify the geometries in the graph representation using the Douglas-Peucker algorithm \cite{douglas1973algorithms} because of its simplicity and robustness. We have also tested the Visvalingam's \cite{visvalingam1993line} on our dataset; however, it does not work as well as the Douglas-Peucker algorithm.


\section{TGGLines approach}\label{sec:approach}

In  this  section,  we  elaborate  the  proposed TGGLines method, including image representation (Section \ref{sec:TGGLines_img_representation}), concepts (Section \ref{sec:TGGLines_concepts}), and workflow illustrated using a simple example (Section \ref{sec:TGGLines_workflow}).

\subsection{TGGLines image representation} \label{sec:TGGLines_img_representation}

The image representation used in our TGGLines is the skeleton graph, which is a topological graph generated from a image skeleton \cite{mlg2019_7}. We use Zhang-Suen thinning algorithm \cite{zhang1984fast} for image skeleton extraction, as it is well-known and robust.

\begin{figure*}[!htb]
\begin{center}
   \begin{subfigure}{0.25\textwidth}
      \includegraphics[width=\textwidth]{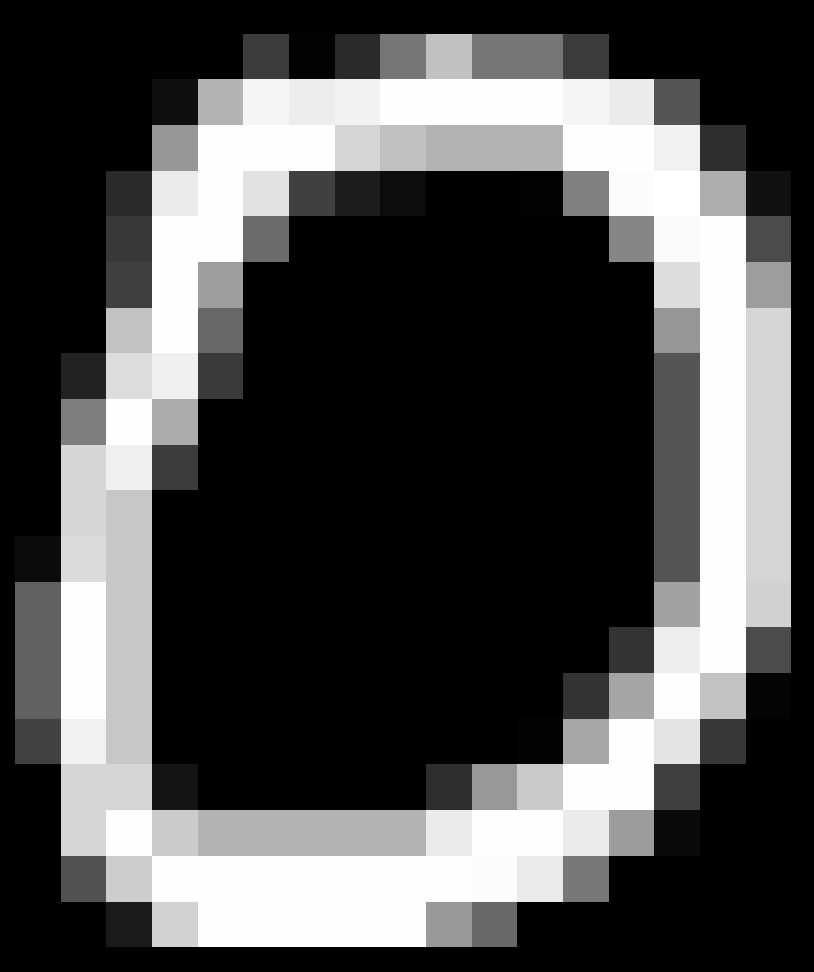}
      \caption{Input image}
    \end{subfigure}
    \hfill
    \begin{subfigure}{0.25\textwidth}
      \includegraphics[width=\textwidth]{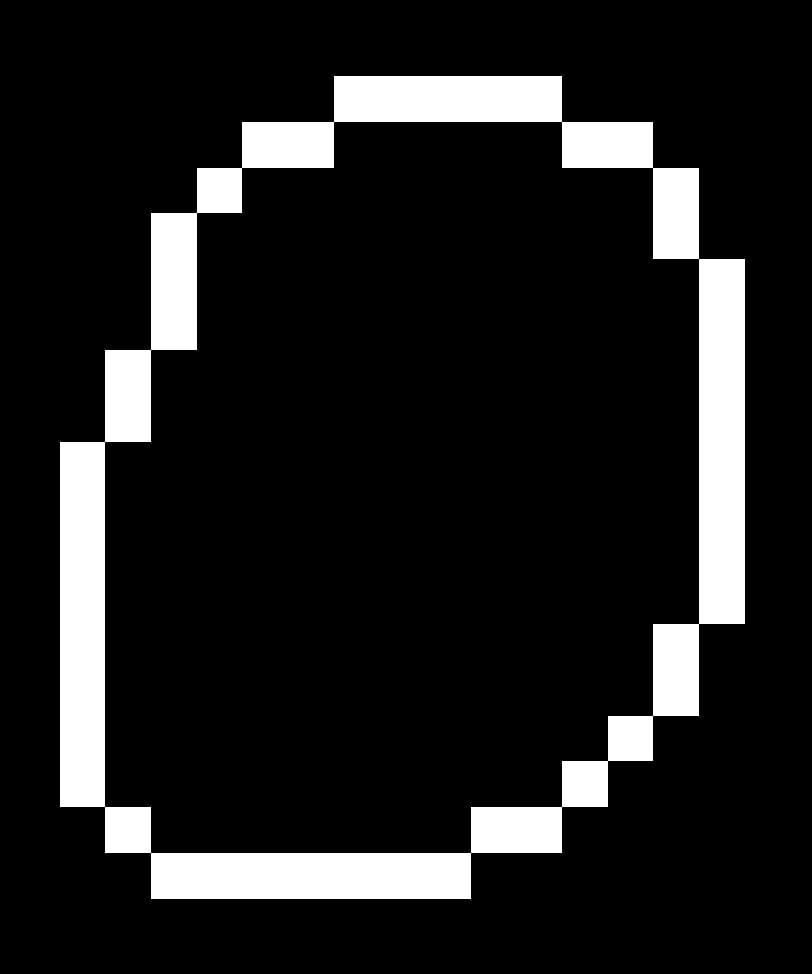}
      \caption{Image skeleton}
    \end{subfigure}
    \hfill
    \begin{subfigure}{0.25\textwidth}
      \includegraphics[width=\textwidth]{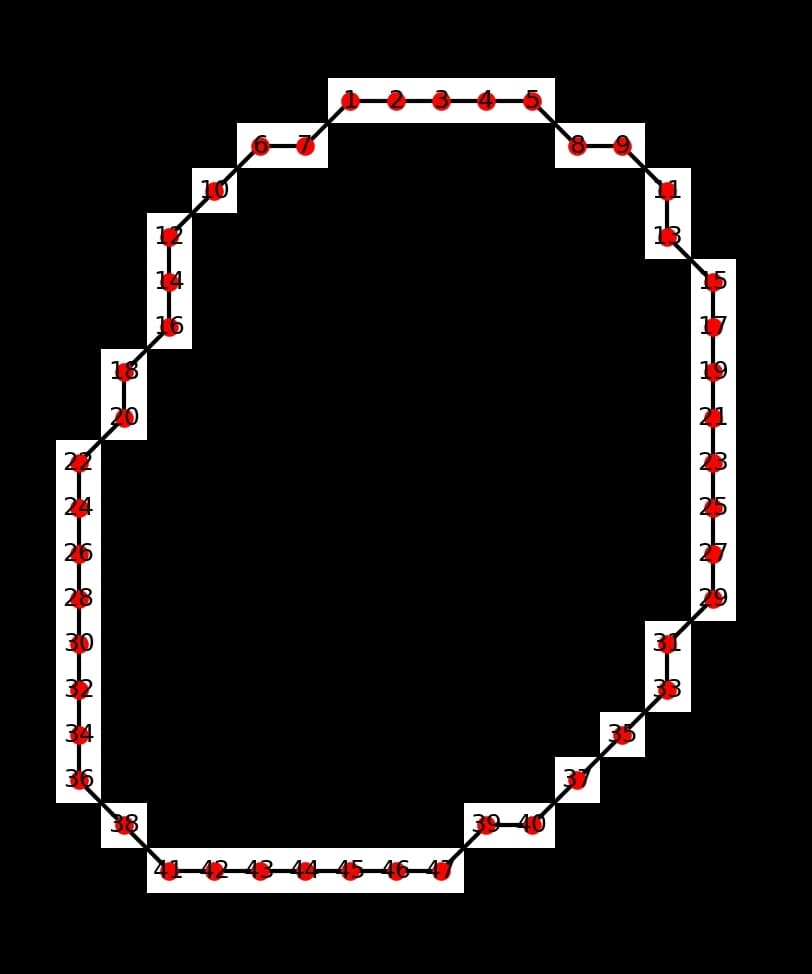}
      \caption{Skeleton graph}
    \end{subfigure}
\end{center}
   \caption{An example of skeleton graph image representation . Figure \ref{fig:skeletonG_img_representation_example} (a) shows the input image. Figure \ref{fig:skeletonG_img_representation_example} (b) shows the image skeleton extracted from the input image. Figure \ref{fig:skeletonG_img_representation_example} (c) provides the skeleton graph corresponding to the skeleton present in (b).}
\label{fig:skeletonG_img_representation_example}
\end{figure*}

See Figure \ref{fig:skeletonG_img_representation_example} below for an illustration of the image representation we use in our TGGLines approach (the handwritten digit image used here is taken from the MNIST dataset\cite{lecun1998gradient}). In the skeleton graph, each node represents a pixel in the image skeleton, and each edge indicates that the two pixels it connects are neighbors.

\subsection{TGGLines concepts} \label{sec:TGGLines_concepts}

\begin{description}[font=\normalfont]
    \item [\textbf{Skeleton graph:}] A skeleton graph is an embedded graph generated from an image skeleton, where each node represents a pixel in the image skeleton, and an edge betwee two pixel nodes indicate the two pixels are neighbors.

    \item [\textbf{Path:}] Each path is an embedded graph that is a subgraph of the skeleton graph, which consists of non-salient nodes segmented by salient nodes (e.g., junction nodes and end nodes, defined in Section \ref{sec:node_type_in_skeletonG}) and edges connecting the nodes. 
     
     \item [\textbf{Line segment connectivity graph (LSCG):}] It is an embedded graph generated from the skeleton graph, where $N$ is a set of nodes each representing a path, and $E$ is a set of edge representing salient nodes (e.g., junction nodes or end nodes). each node will have an attribute that pointing to its corresponding path it represents. LSCG is used to organized segmented paths.

    \item [\textbf{Simplified LSCG:}] It is the LSCG that the paths it organize are simplified.

\end{description}

\subsubsection{Node type in skeleton graph} \label{sec:node_type_in_skeletonG}
There are three types of salient nodes in our TGGLines. We will use them to segment a skeleton graph to multiple paths for simplification.

\begin{itemize}
    \item \textbf{End node:} A (pixel) node that has only 1 neighbor.
    \item \textbf{Junction node:} A (pixel) node that has $n$ neighbors where n $>$ 2 
    \item \textbf{Turning node:} A (pixel) node that has two neighbors.
\end{itemize}

\subsection{TGGLines workflow}  \label{sec:TGGLines_workflow}
The TGGLines workflow is presented in Figure \ref{fig:TGGLines_workflow_illustrativeexample} below visually illustrated by a simple and straightforward example.

\begin{figure*}[!htb]
\begin{center}
   \begin{subfigure}{0.3\textwidth}
      \includegraphics[width=\textwidth]{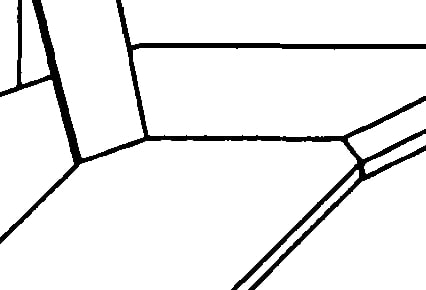}
      \caption{Input diagram image}
    \end{subfigure}
    \hfill
    \begin{subfigure}{0.3\textwidth}
      \includegraphics[width=\textwidth]{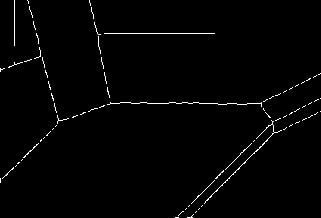}
      \caption{Image skeleton extraction}
    \end{subfigure}
    \hfill
    \begin{subfigure}{0.32\textwidth}
      \includegraphics[width=\textwidth]{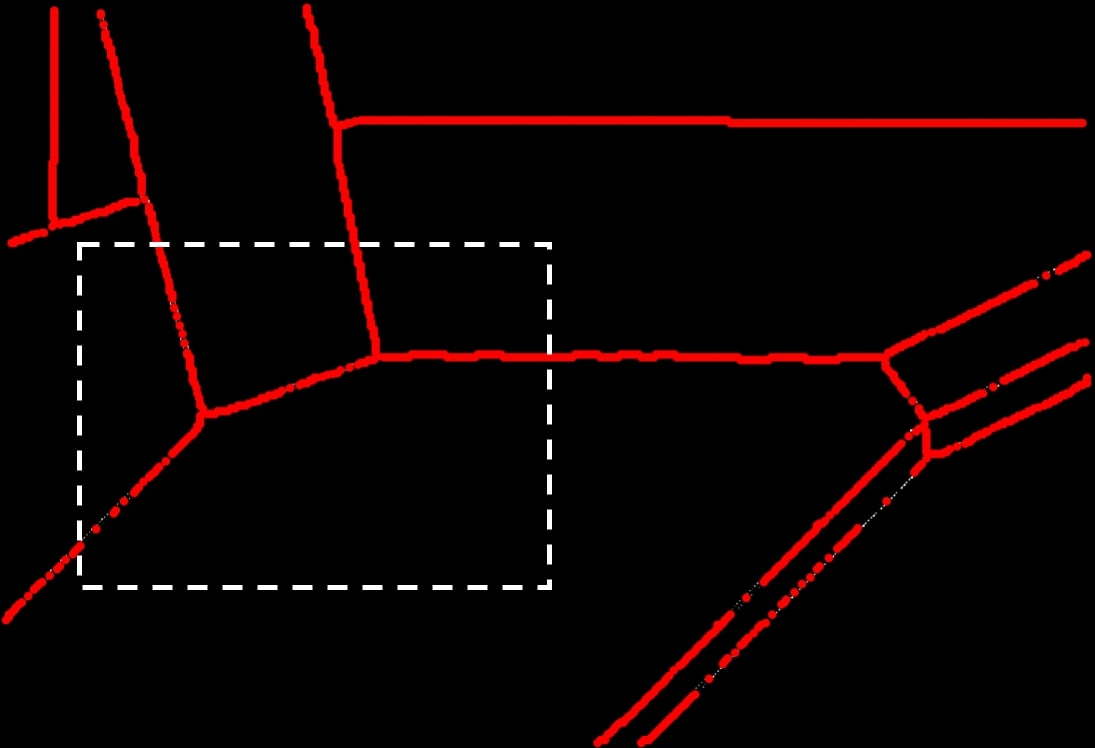}
      \caption{Skeleton graph: a topological graph generated from skeleton \cite{mlg2019_7}}
    \end{subfigure}
    \hfill
    \begin{subfigure}{0.32\textwidth}
      \includegraphics[width=\textwidth]{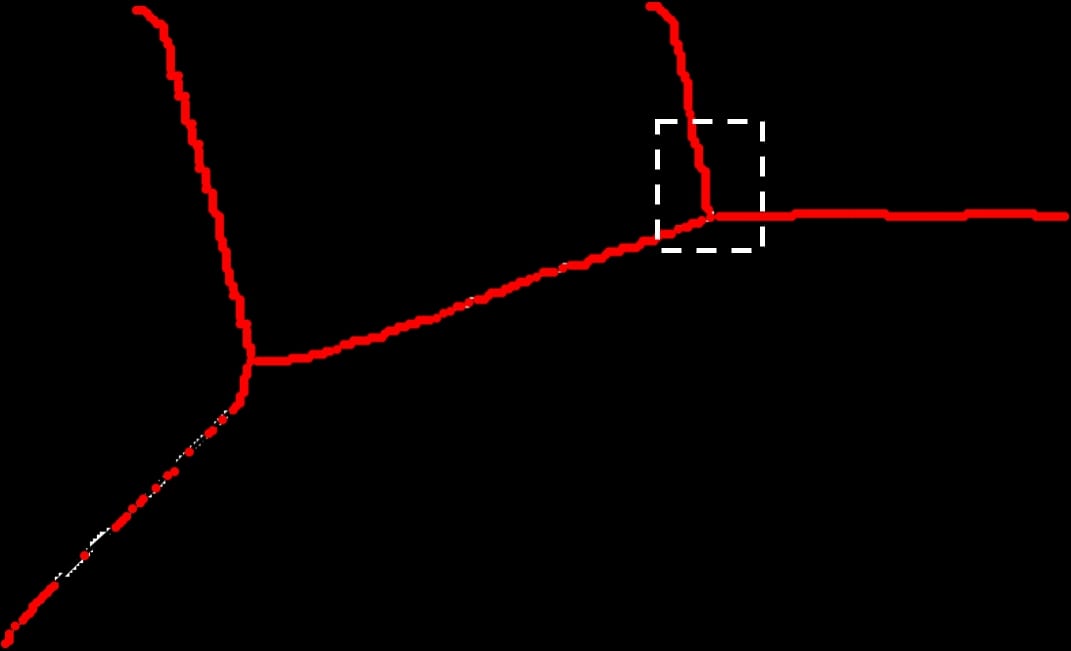}
      \caption{Skeleton graph (zoomed-in detail)}
    \end{subfigure}
    \hfill
    \begin{subfigure}{0.32\textwidth}
      \includegraphics[width=\textwidth]{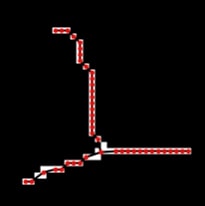}
      \caption{Skeleton graph (zoomed-in detail)}
    \end{subfigure}
    \hfill    
    \begin{subfigure}{0.33\textwidth}
      \includegraphics[width=\textwidth]{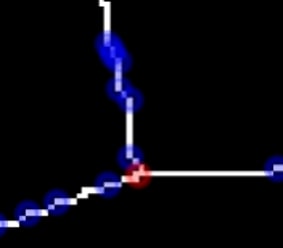}
      \caption{Detecting salient nodes (zoom-in detail)}
    \end{subfigure}
    \hfill
    \begin{subfigure}{0.2\textwidth}
      \includegraphics[width=\textwidth]{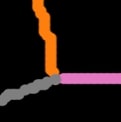}
      \caption{Segmenting skeleton graph to paths and generating LSCG}
    \end{subfigure}
    \hfill
    \begin{subfigure}{0.32\textwidth}
      \includegraphics[width=\textwidth]{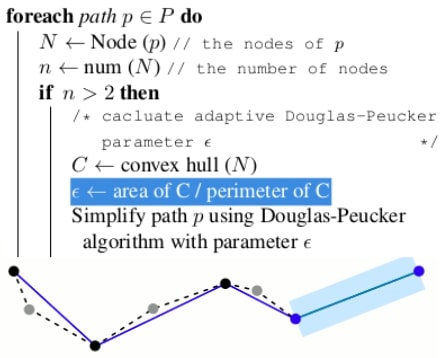}
      \caption{Simplifying paths organized by LSCG (adaptive parameter)}
    \end{subfigure}
    \hfill
    \begin{subfigure}{0.33\textwidth}
      \includegraphics[width=\textwidth]{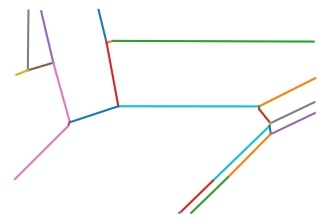}
      \caption{Detected line segments (organized by LSCG)}
    \end{subfigure}
    
\end{center}
   \caption{TGGLines workflow illustrated with a simple and straightforward example. Note that in (b) it is inverted for image skeleton extraction. In (c), the red node indicates junction nodes and blue nodes indicate turning node (definition of salient point types can be found in Section \ref{sec:node_type_in_skeletonG}). in (g) LSCG represents line segment connectivity graph. Each node in LSCG represents a path, for example in (g) there are three different paths.} 
\label{fig:TGGLines_workflow_illustrativeexample}
\end{figure*}

\section{TGGLines algorithms}\label{sec:tgg_algs}

In  this  section,  we provide the algorithms we developed for the TGGLines introduced in Section \ref{sec:approach} above.


The overview pseducode for the TGGLines algorithm is provided in Algorithm \ref{alg:TGGLines_alg}.

We first extract skeleton from an input diagram image. then a skeleton graph is generated from the skeleton, after that, salient points (defined in Section \ref{sec:node_type_in_skeletonG}) are detected by counting the incident edges for each pixel node in the skeleton graph.
Then the skeleton graph is segmented to multiple paths using the detected salient nodes; meanwhile a line segment connected graph is generated while segmenting skeleton graph to paths. Then we simplyfing paths oragzed in LSCG using the Douglas-Peucker algorithm \cite{douglas1973algorithms}, detailed in Algorithm \ref{alg:simplification}.

\begin{algorithm}
\DontPrintSemicolon

    \KwInput{A diagram image $I$}
    \KwOutput{An embedded graph $G_{lsc} = \{N, E\}$ representing the simplified line segment connectivity graph from $I$ \tcp*{the nodes of an embedded graph contain coordinates, therefore the graph can be drawn uniquely on a plane}}
    
    $S \gets $ skeleton($I$) \;
    
     $G_s \gets $ skeletonGraph($S$) \;

     \tcc{salient nodes detection for segmenting line segments}      
      $N_s \gets $ salientNodes ($G_s$) \; \tcp*{detecting salient nodes}
      $P \gets $ Paths($G_s$,$N_s$) \;\tcp*{segmenting paths}

      $G_{lsc} \gets $ lineSegConnectvityGraph ($G_s$, $N_s$) \;
          
      $G_{slsc} \gets $ simplifyingLineSegConnectvityGraph ($G_{lsc}$) \;
       return  $G_{slsc}$ \;
 \caption{TGGLines algorithm}\label{alg:TGGLines_alg}
\end{algorithm}

\begin{figure}[!htb]
\begin{center}
   \includegraphics[width=0.95\linewidth]{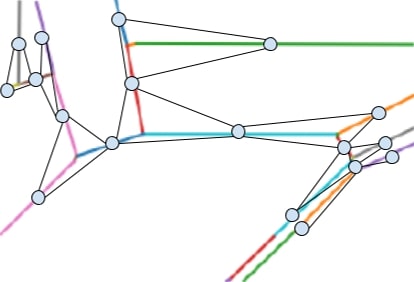}
\end{center}
   \caption{Line segment connectivity graph (LSCG) for the example shown in the workflow Figure \ref{fig:TGGLines_workflow_illustrativeexample} above. each light blue node represents a path, if two paths are connected by a junction node or turning node, there is an edge between the nodes.}
\label{fig:LSCG_for_TGGLines_workflow_example}
\end{figure}

The LSCG for the illustrative example shown in the workflow Figure \ref{fig:TGGLines_workflow_illustrativeexample} is given in Figure \ref{fig:LSCG_for_TGGLines_workflow_example}.

 \begin{algorithm}
\DontPrintSemicolon

    \KwInput{Segmented paths $P$ organized by LSCG }
    \KwOutput{An updated LSCG pointing to simplified paths $S$}

     $S \gets null$ \tcp*{Initialization}   
    \ForEach {path $p \in  P$}{  
        $N \gets $ Node ($p$) \tcp{the nodes of $p$} 
        $n \gets $ num ($N$) \tcp{the number of nodes} 
        \If{ $n > 2$}{
            \tcc{cacluate adaptive Douglas-Peucker parameter $\epsilon$}
             $C \gets $ convex hull ($N$)   \;
            $\epsilon \gets $ area of C  / perimeter of C  \;
            Simplify path $p$ using Douglas-Peucker algorithm with parameter $\epsilon$\;
            $new\_path \gets$ simplified $p$ \;
            
            append $new\_path$ to $S$ \;
        }
    }
       return $S$, and update $LSCG$ \;

 \caption{Simplifying paths organized by LSCG} \label{alg:simplification}
\end{algorithm}

\section{Experiments and evaluation} \label{sec:experiments_evaluation}

In this section, we provide details about dataset (Section \ref{sec:dataset}), experiments and results (Section \ref{sec:experiments_results}), qualitative (Section \ref{sec:evaluation_qualitative}) and quantitative (Section \ref{sec:evaluation_quantitative}) evaluation.

\subsection{Dataset}\label{sec:dataset}
To evaluate TGGLines, and compare with state-of-the-art methods, we develop a simple interface to manually annotate 10 diagram images taken from the 2000 Binary Patent Image Database developed by Multimedia Knowledge and Social Media Analytics Laboratory (MKLab)  \cite{mklab2010}. The database contains images from patents maintained by the European Patent Office.

\begin{figure}[!htb]
\begin{center}
   \begin{subfigure}{0.24\textwidth}
      \includegraphics[width=\textwidth]{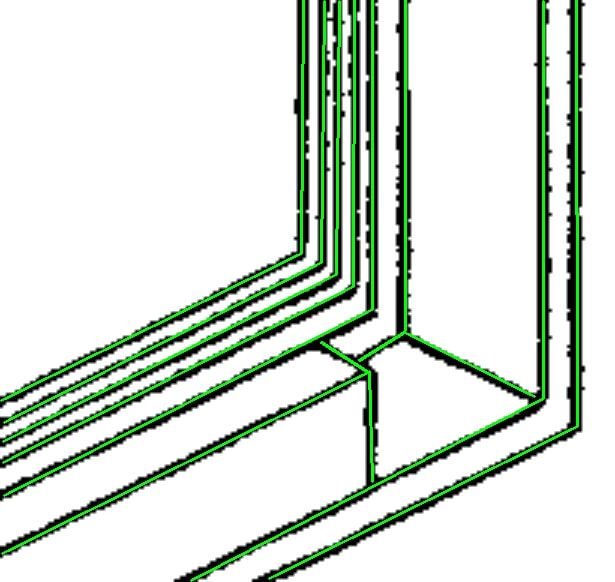}
      \caption{}
    \end{subfigure}
    \hfill
    \begin{subfigure}{0.22\textwidth}
      \includegraphics[width=\textwidth]{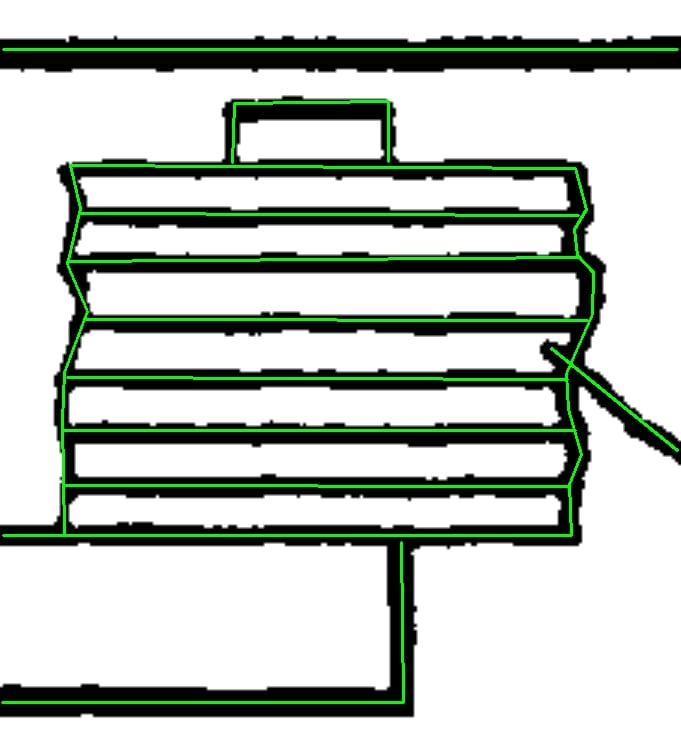}
      \caption{}
    \end{subfigure}
\end{center}
   \caption{Annotation examples: Line segments of (a) varying angles and spacing intervals; and (b) even interval. Examples have different resolution.} 
\label{fig:lineseg_annotation_examples}
\end{figure}

For readability, we take partial images from 10 selected samples (cropped without changing the resolution of the original images), the file size ranges from 53 KB (212x171, \#3 in Table \ref{tab:result_images_part1}) to 307 KB (524x566, \#7 in Table \ref{tab:result_images_part2}). Our dataset is small yet representative. Samples are carefully selected from the 2000 patent images, by considering this selection criteria: (1) different complexity of line directions (e.g., vertical, horizontal, other arbitrary angles), (2) spacing between line segments (roughly equal spacing, sparse, dense), (3) topological relations (e.g., single line segment, intersection, turning, circular). Figure \ref{fig:lineseg_annotation_examples} shows two annotated examples. 
The images in the experiment, image \#01 to \#10, are the partial images taken from the following images from the patent image dataset \cite{mklab2010}: 01779, 01126, 00501, 00780, 00971, 00575, 00429, 01267, 00811, and 01140.

\subsection{Experiments and results}\label{sec:experiments_results}

We implement TGGLines in Python using OpenCV, Scikit-image \cite{van2014scikit}, SymPy \cite{meurer2017sympy}, and NetworkX \cite{hagberg2008exploring}. We compare TGGLines with state-of-the-art methods: PPHT \cite{matas2000robust}, LSD \cite{von2010lsd}, EDLines \cite{akinlar2011edlines} and Linelet \cite{cho2017novel}. Results can be found in the sets of figures organized in Tables \ref{tab:result_images_part1} and \ref{tab:result_images_part2}. Parameter settings (Table \ref{tab:parameter_settings}) for each method are provided in Appendix \ref{app:parameter_setting_computational_time}, and the computing environment details can be found in Appendix \ref{app:computing_env}.

\begin{table*}
  [!htb] \caption{Line segment detection results on images \#01 --- \#05 of the dataset.} \label{tab:result_images_part1}
\setlength\tabcolsep{3pt} 
\centering
\begin{tabular}{|m{0.9cm}<{\centering}|m{2cm}<{\centering}|m{2cm}<{\centering}|m{2cm}<{\centering}|m{2cm}<{\centering}|m{2cm}<{\centering}|m{2cm}<{\centering}|m{2cm}<{\centering}|}
\hline
      Img \# & Input images & Ground truth & PPHT & LSD & EDLines & Linelet & \textbf{TGGLines}\\
      \hline 
      
      01 & 
      \includegraphics[scale = 0.5]{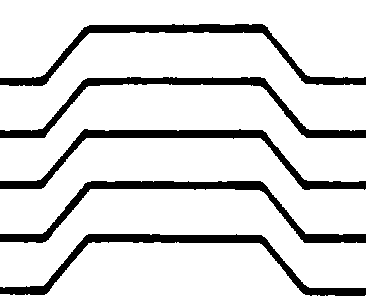} &  
      \includegraphics[scale = 0.04]{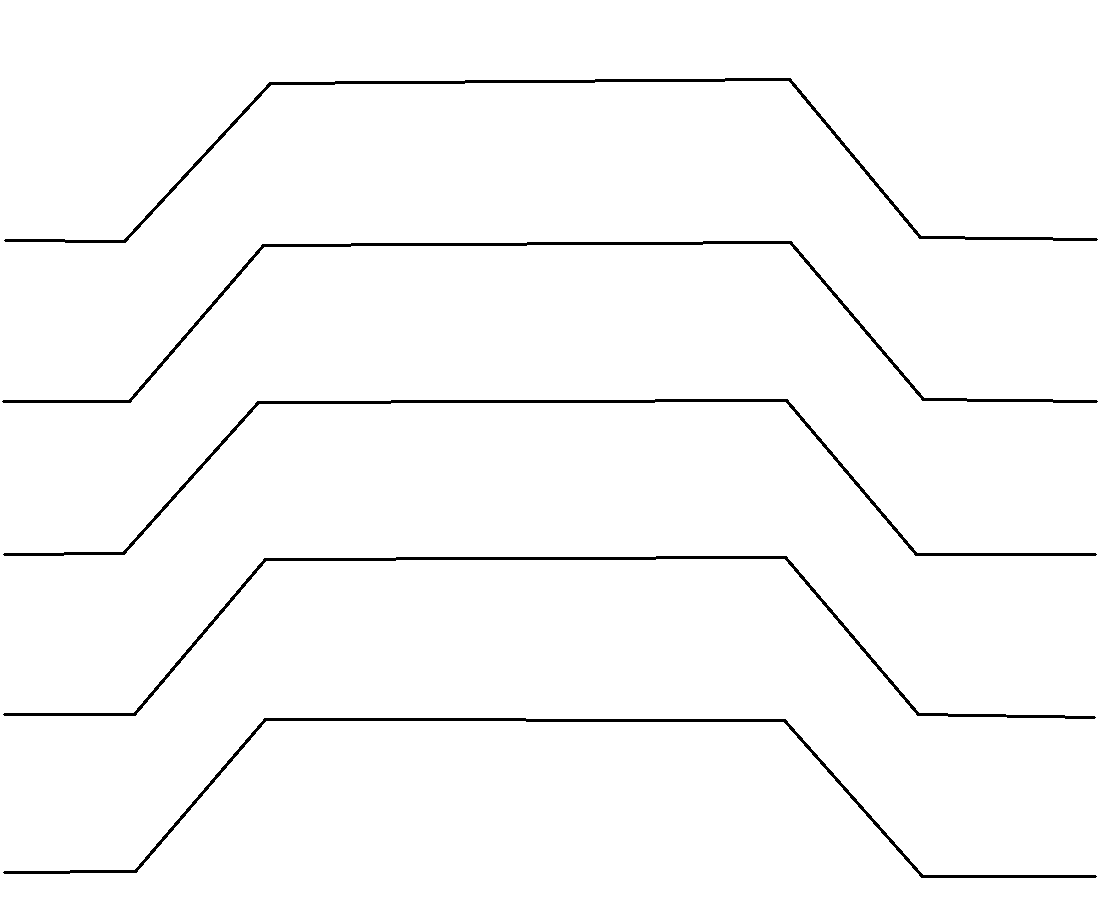}&  
      \includegraphics[scale = 0.215]{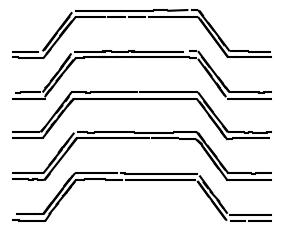}&  
      \includegraphics[scale = 0.12]{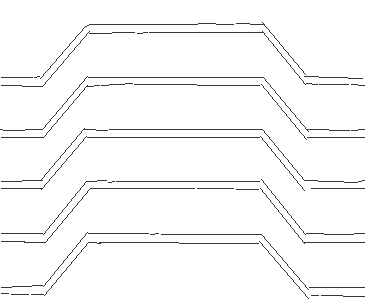}&  
      \includegraphics[scale = 0.12]{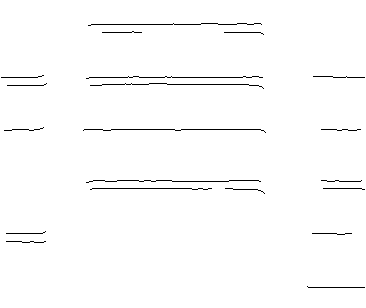}&  
      \includegraphics[scale = 0.17]{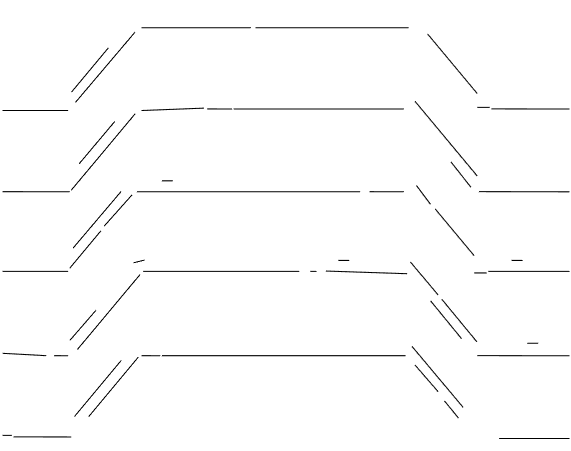}&   
      \includegraphics[scale = 0.21]{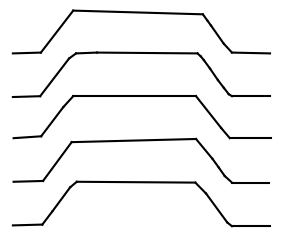}\\
      \hline
      
      02 & 
      \includegraphics[scale = 0.5]{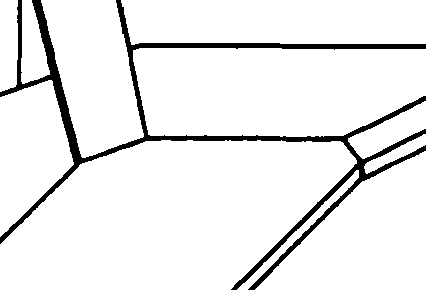}&  
      \includegraphics[scale = 0.04]{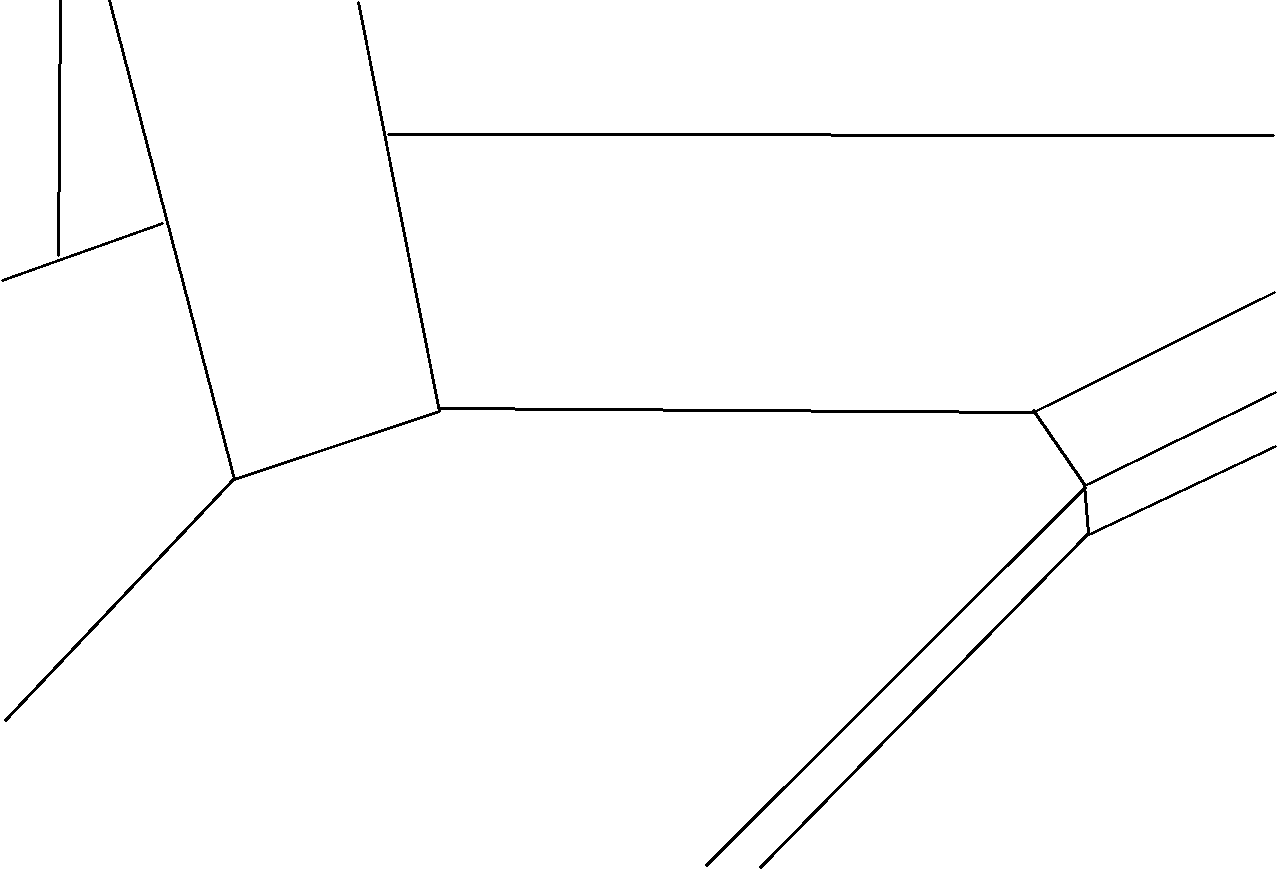}&  
      \includegraphics[scale = 0.23]{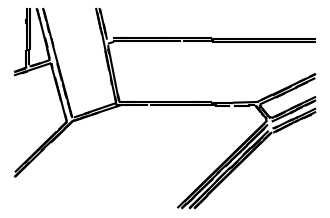}&  
      \includegraphics[scale = 0.12]{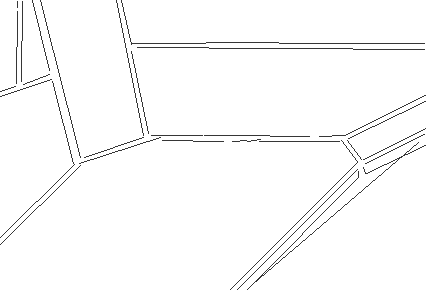}&  
      \includegraphics[scale = 0.12]{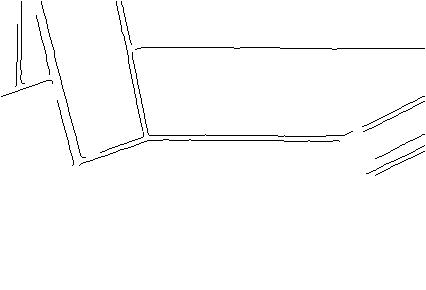}&  
      \includegraphics[scale = 0.15]{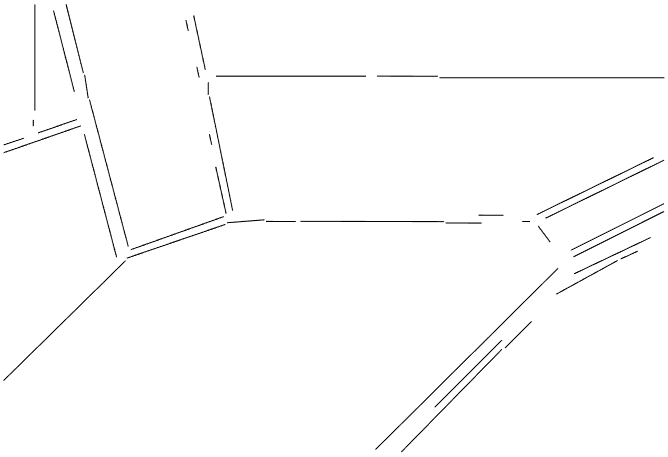}&   
      \includegraphics[scale = 0.22]{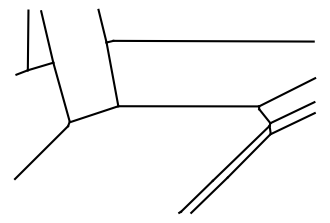}\\
      \hline
      
      03 & 
      \includegraphics[scale = 0.89]{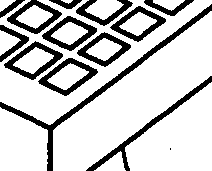}&  
      \includegraphics[scale = 0.058]{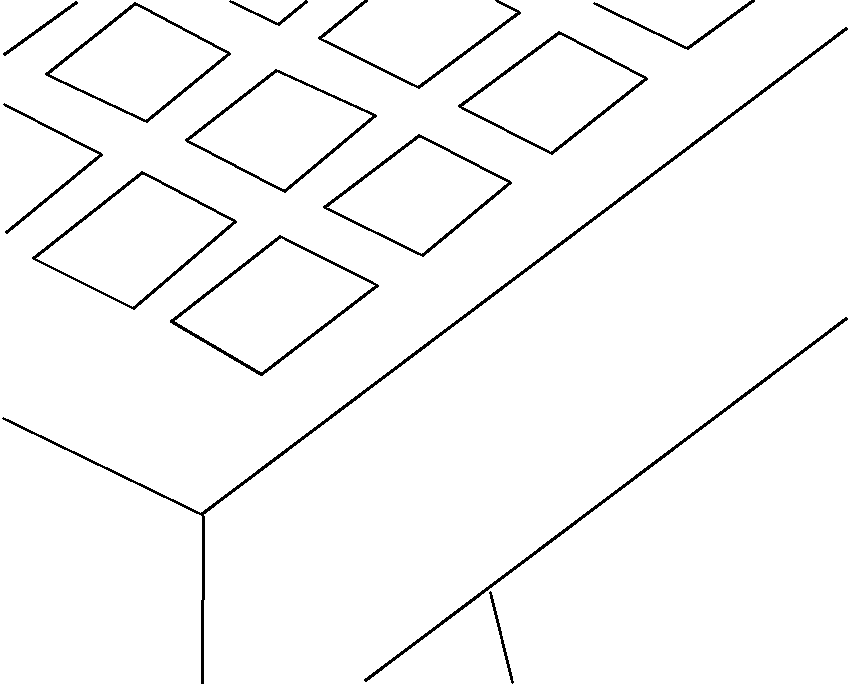}&  
      \includegraphics[scale = 0.42]{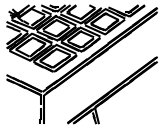}&  
      \includegraphics[scale = 0.25]{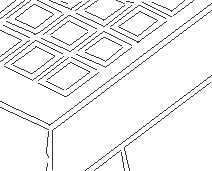}&  
      \includegraphics[scale = 0.3]{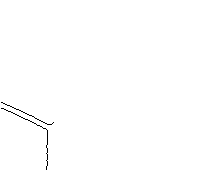}&  
      \includegraphics[scale = 0.35]{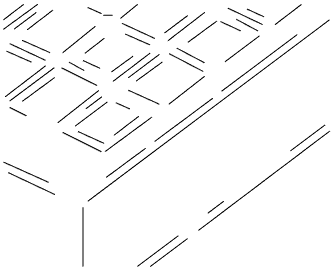}&   
      \includegraphics[scale = 0.38]{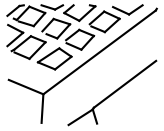}\\
      \hline
      
      04 & 
      \includegraphics[scale = 0.5]{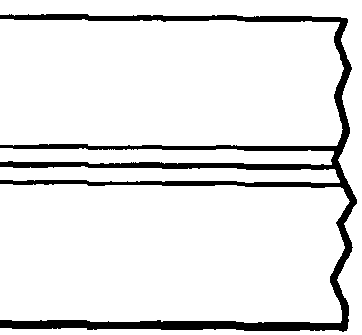}&  
      \includegraphics[scale = 0.06]{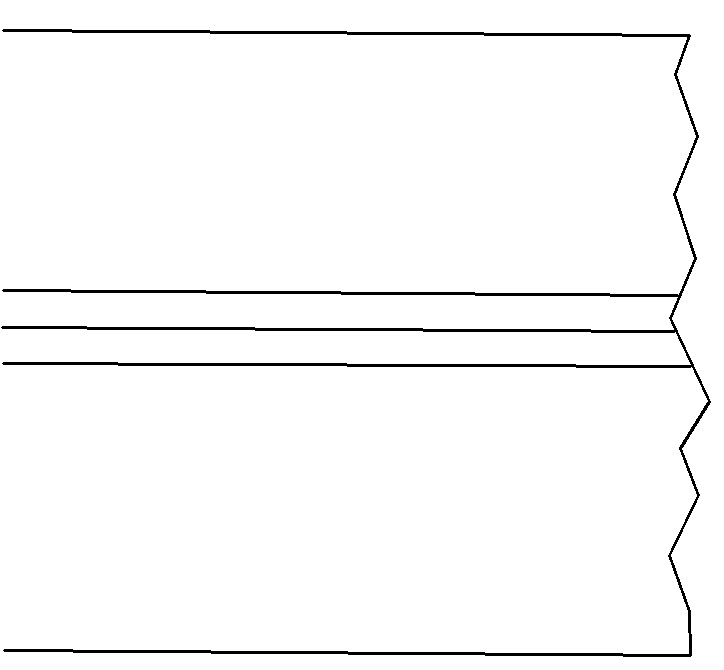}&  
      \includegraphics[scale = 0.23]{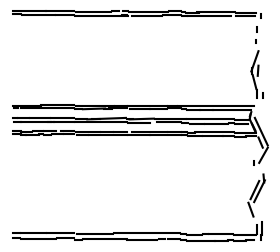}&  
      \includegraphics[scale = 0.12]{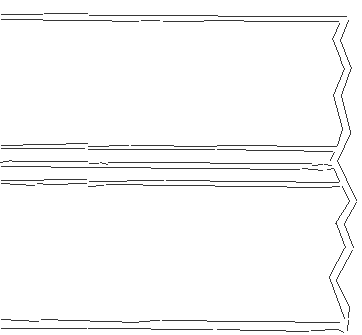}&  
      \includegraphics[scale = 0.12]{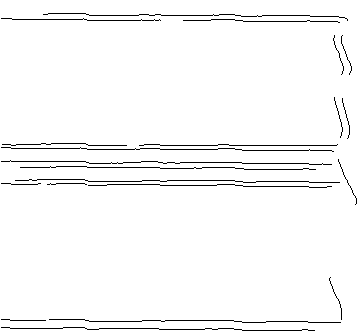}&  
      \includegraphics[scale = 0.17]{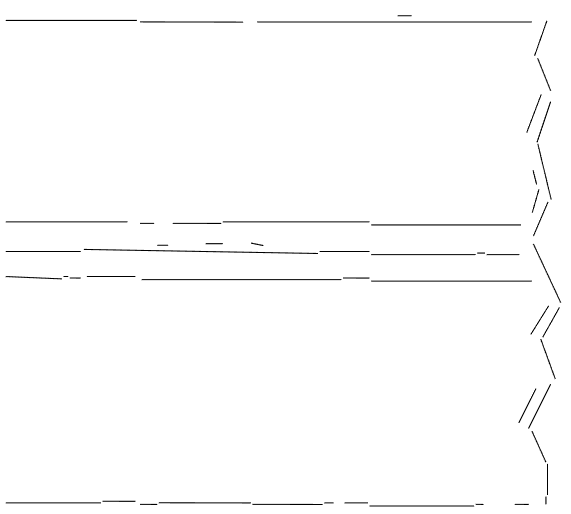}&   
      \includegraphics[scale = 0.22]{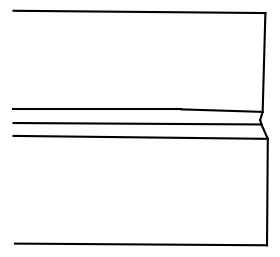}\\
      \hline
      
      05 & 
      \includegraphics[scale = 0.7]{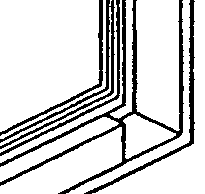}&  
      \includegraphics[scale = 0.055]{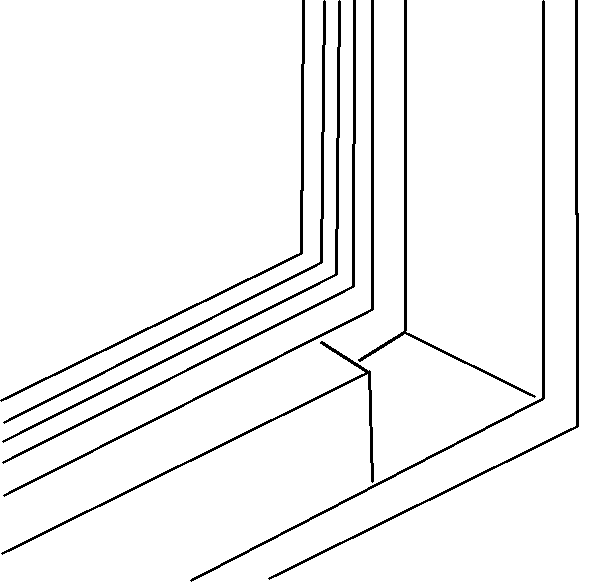}&  
      \includegraphics[scale = 0.33]{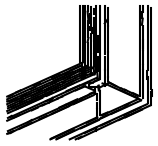}&  
      \includegraphics[scale = 0.17]{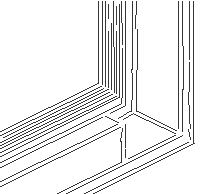}&  
      \includegraphics[scale = 0.17]{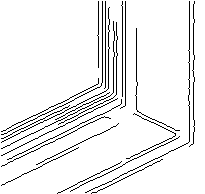}&  
      \includegraphics[scale = 0.22]{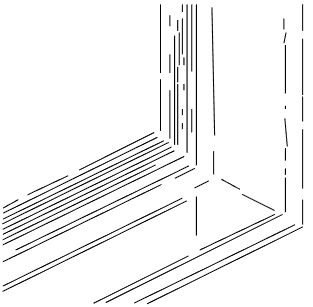}&   
      \includegraphics[scale = 0.29]{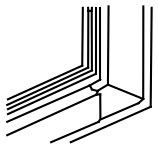}\\
      \hline
  \end{tabular}
\end{table*}

\begin{table*}
  [!htb] \caption{Line segment detection results on images \#06 --- \#10 of the dataset.} \label{tab:result_images_part2}
  \setlength\tabcolsep{1pt} 
  \centering
  \begin{tabular}{|m{0.9cm}<{\centering}|m{2.2cm}<{\centering}|m{2.2cm}<{\centering}|m{2.2cm}<{\centering}|m{2.2cm}<{\centering}|m{2.2cm}<{\centering}|m{2.2cm}<{\centering}|m{2.2cm}<{\centering}|}
   \hline 
      Img \# & Input images & Ground truth & PPHT & LSD & EDLines & Linelet & \textbf{TGGLines} \\
      \hline 
      
      06 & 
      \includegraphics[scale = 0.5]{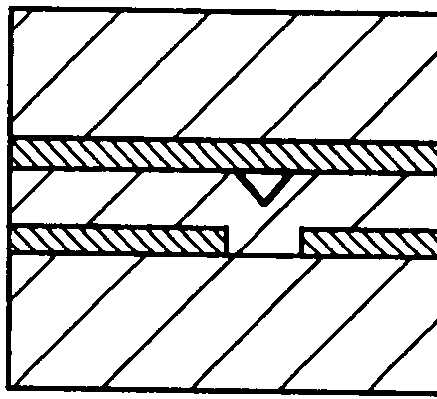}&  
      \includegraphics[scale = 0.065]{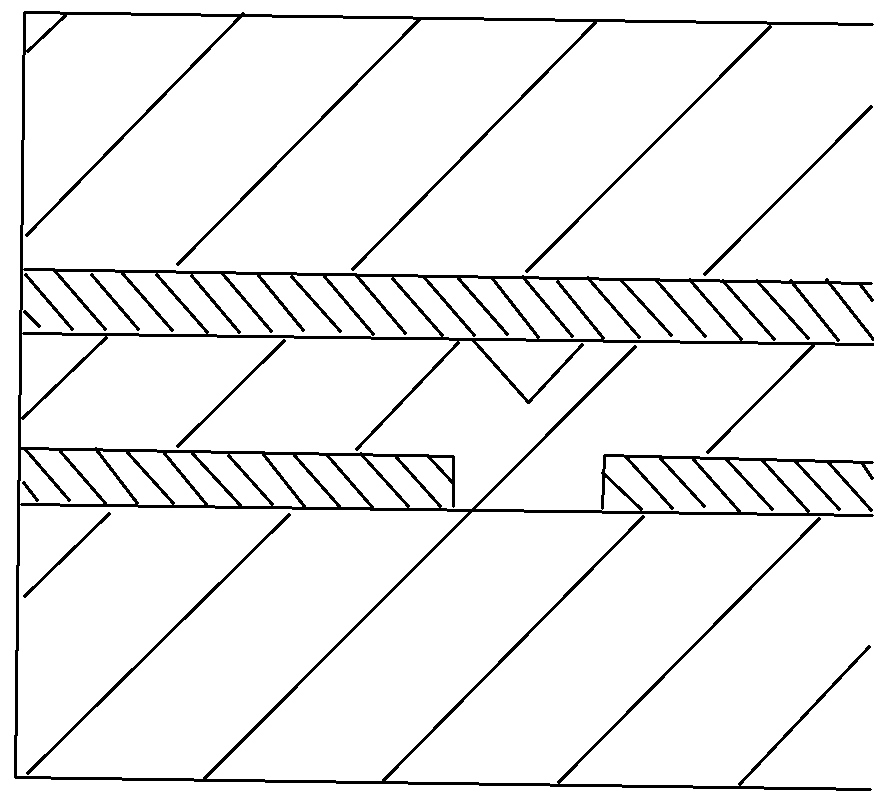}&  
      \includegraphics[scale = 0.25]{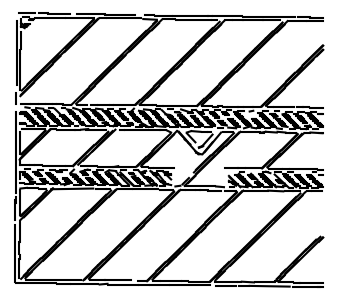}&  
      \includegraphics[scale = 0.12]{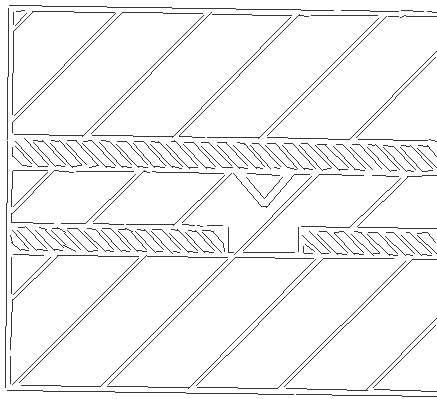}&  
      \includegraphics[scale = 0.12]{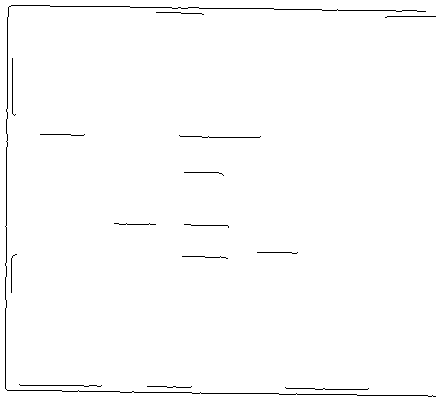}&  
      \includegraphics[scale = 0.08]{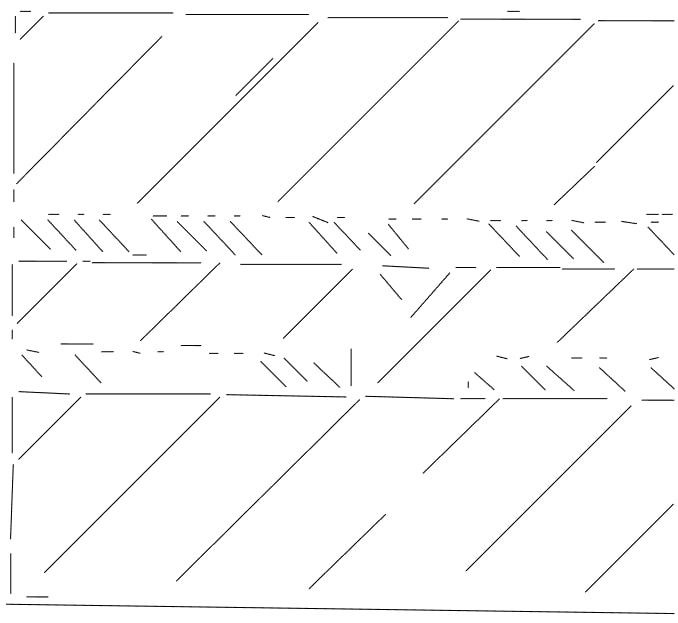}&   
      \includegraphics[scale = 0.25]{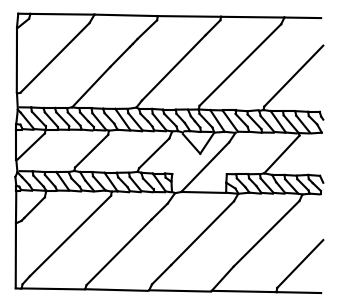}\\
      \hline
      
      07 & 
      \includegraphics[scale = 0.45]{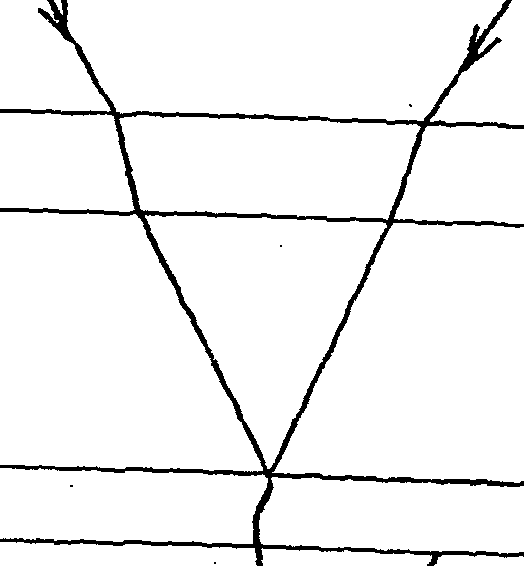}&  
      \includegraphics[scale = 0.095]{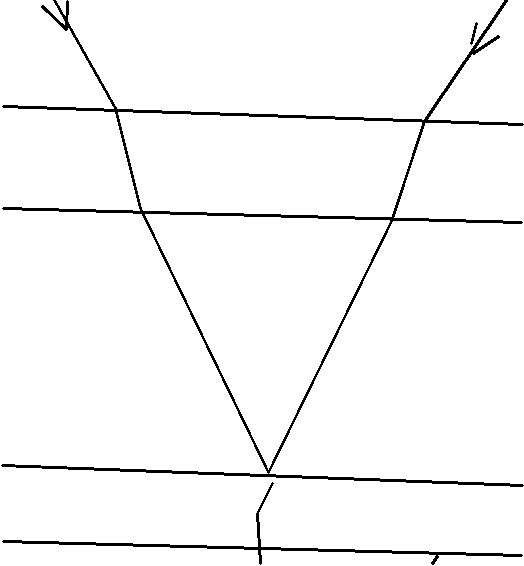}&  
      \includegraphics[scale = 0.2]{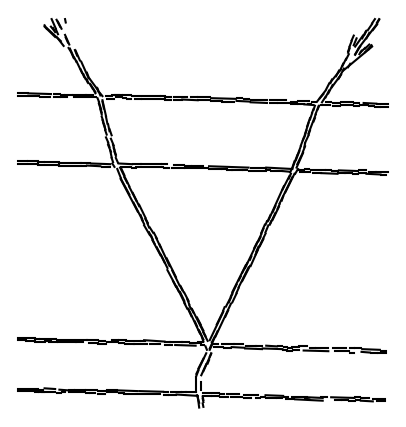}&  
      \includegraphics[scale = 0.1]{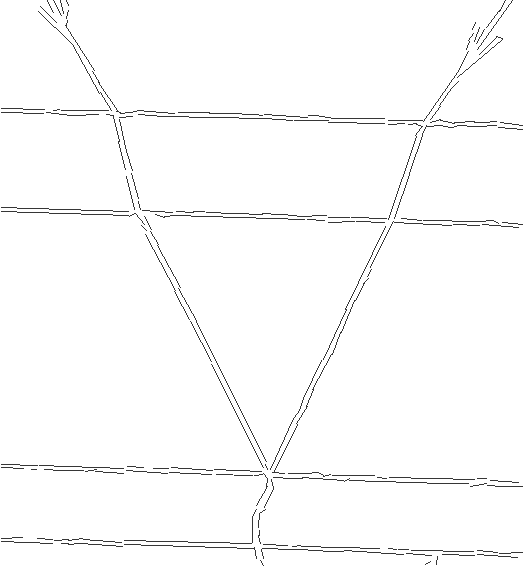}&  
      \includegraphics[scale = 0.12]{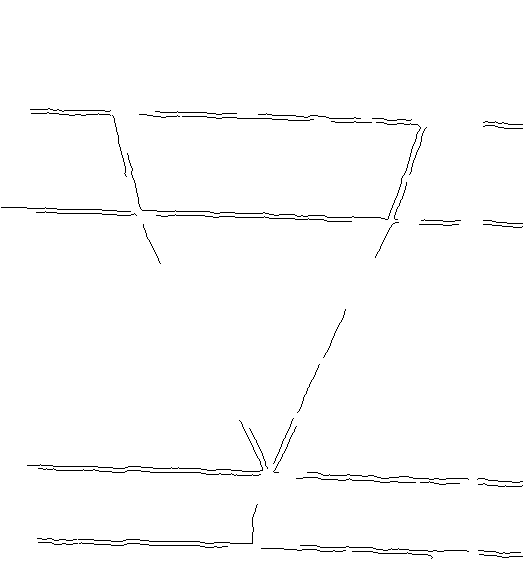}&  
      \includegraphics[scale = 0.07]{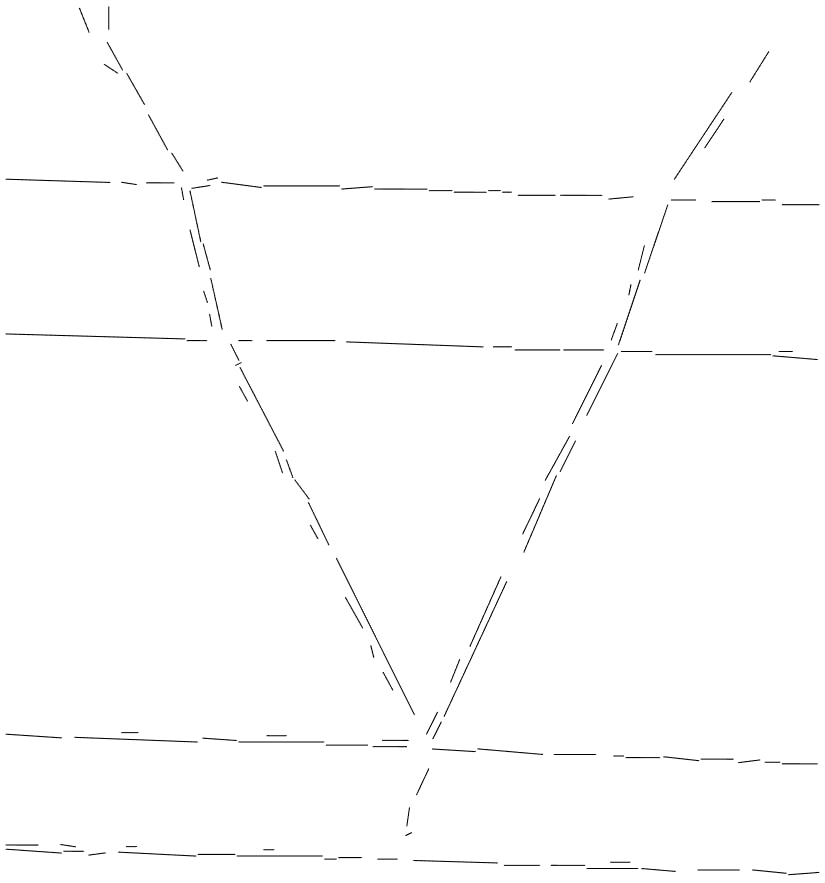}&   
      \includegraphics[scale = 0.21]{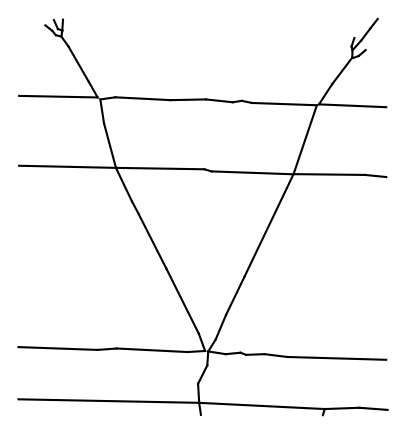}\\
      \hline
      
      08 & 
      \includegraphics[scale = 1.1]{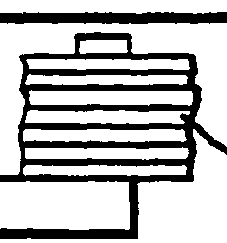}&  
      \includegraphics[scale = 0.085]{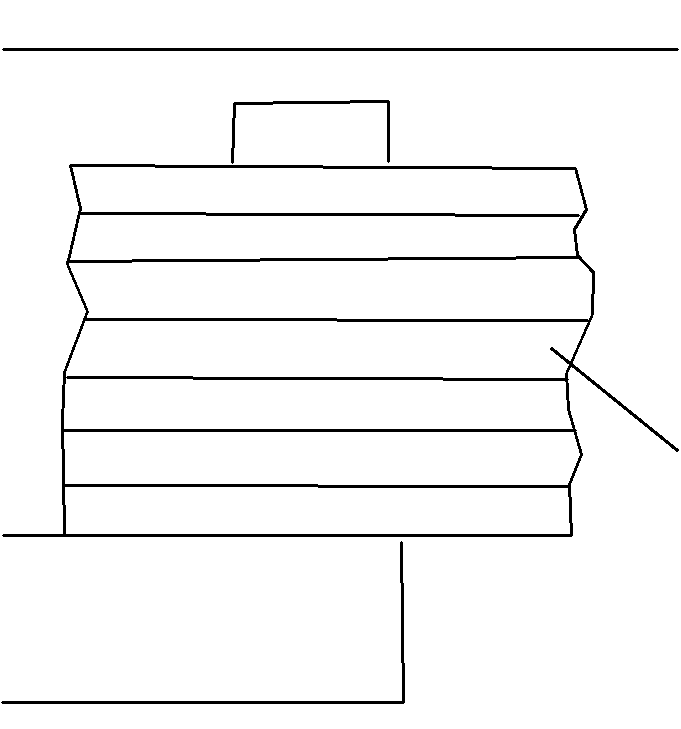}&  
      \includegraphics[scale = 0.45]{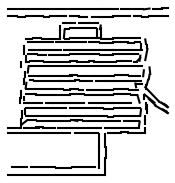}&  
      \includegraphics[scale = 0.25]{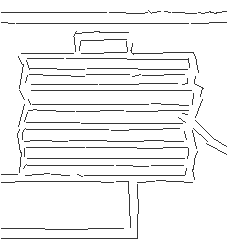}&  
      \includegraphics[scale = 0.25]{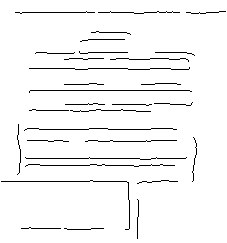}&  
      \includegraphics[scale = 0.35]{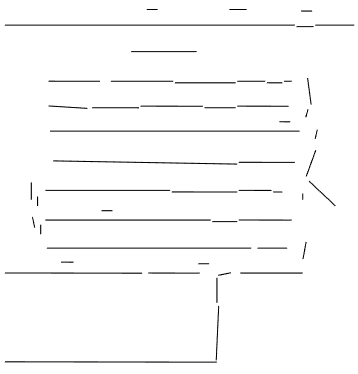}&   
      \includegraphics[scale = 0.42]{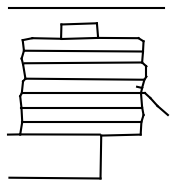}\\
      \hline
      
      09 & 
      \includegraphics[scale = 0.6]{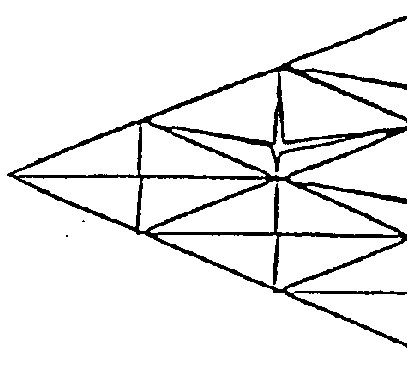}&  
      \includegraphics[scale = 0.07]{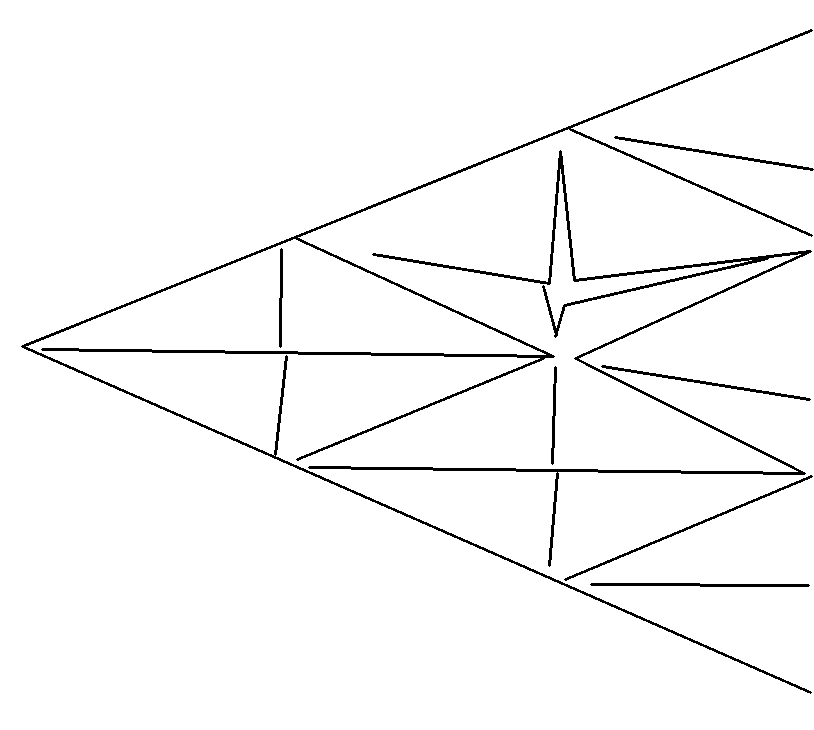}&  
      \includegraphics[scale = 0.28]{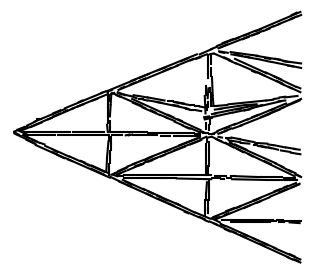}&  
      \includegraphics[scale = 0.15]{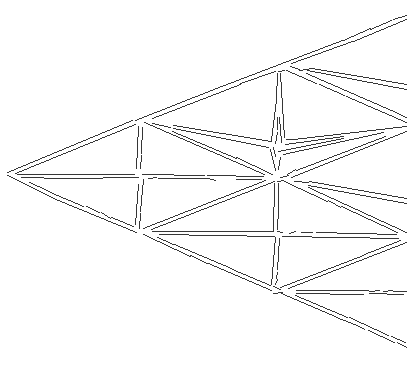}&  
      \includegraphics[scale = 0.15]{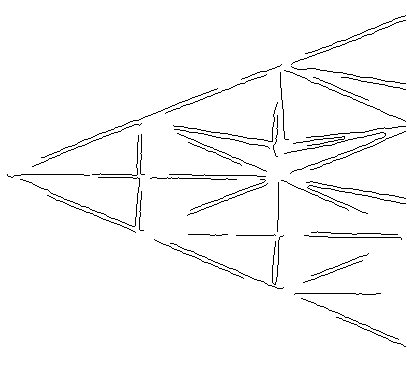}&  
      \includegraphics[scale = 0.2]{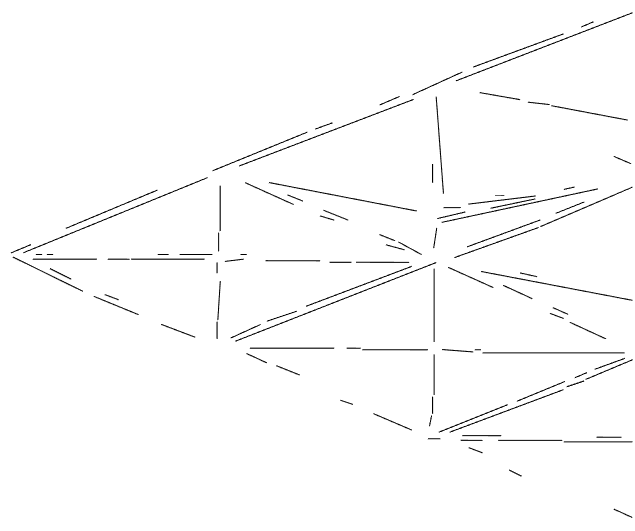}&   
      \includegraphics[scale = 0.25]{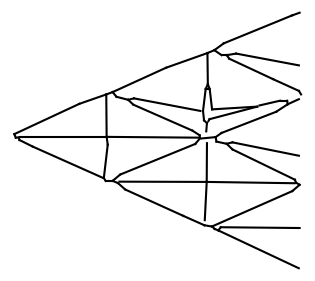}\\
      \hline
      
      10 & 
      \includegraphics[scale = 0.6]{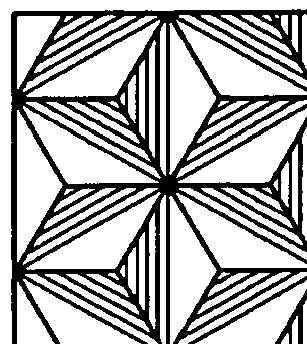}&  
      \includegraphics[scale = 0.075]{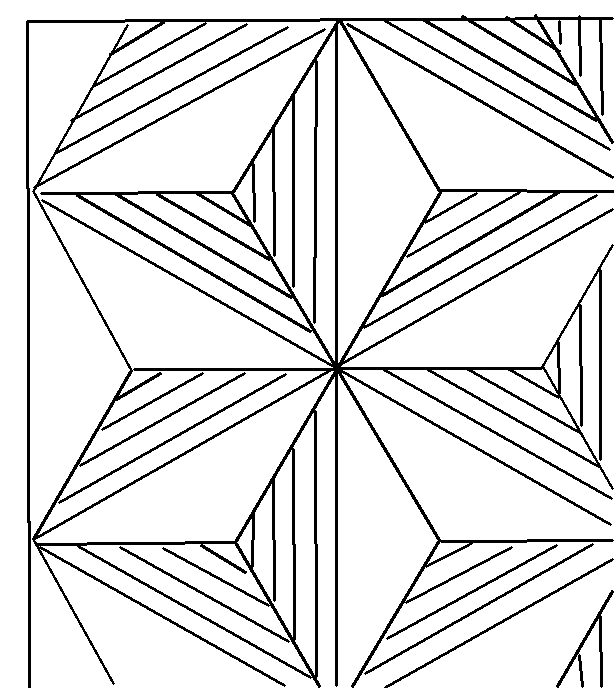}&  
      \includegraphics[scale = 0.28]{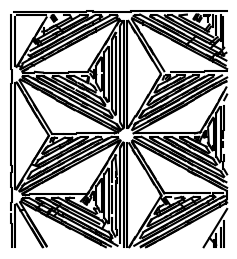}&  
      \includegraphics[scale = 0.15]{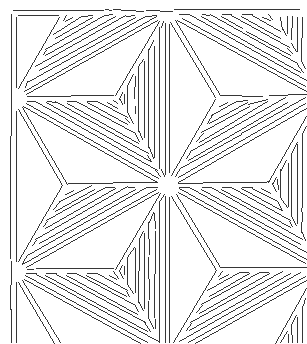}&  
      \includegraphics[scale = 0.15]{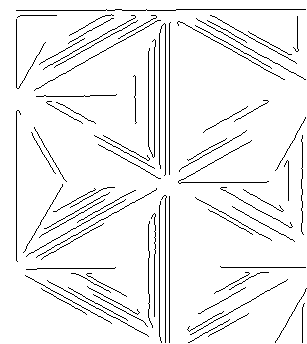}&  
      \includegraphics[scale = 0.2]{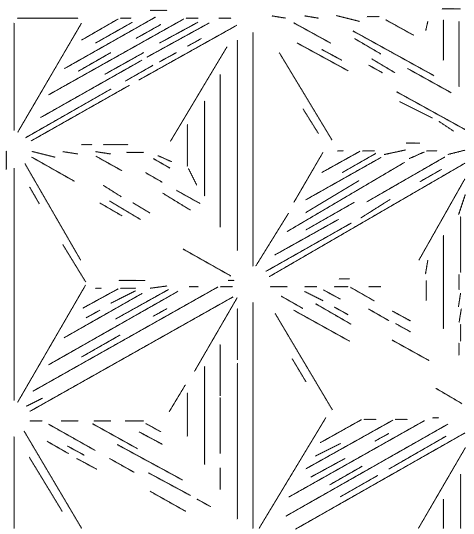}&   
      \includegraphics[scale = 0.28]{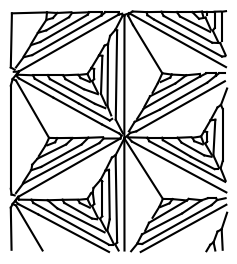}\\
      \hline
  \end{tabular}
\end{table*}

\subsection{Qualitative comparison}\label{sec:evaluation_qualitative}
We qualitatively compare the line segment detection methods as presented in the results Tables \ref{tab:result_images_part1} and \ref{tab:result_images_part2}. Clearly, PPHT detects many tiny line segments and double lines for each true line segment, as it is an edge-based method. LSD and EDLines detect double lines for each line segment in the ground truth; this is not surprising, as they are local gradient-based methods. 

Visually, EDLines performs the worst for our experiment dataset, many line segments cannot be detected by EDLines (especially for image \#3 and \#6 in Table \ref{tab:result_images_part1} and \ref{tab:result_images_part2}, respectively).
LSD detect many tiny lines, similar to PPHT, especially in image \#7 and \#8 in Table \ref{tab:result_images_part2}.

The performance of Linelet is not consistent. For some images, double lines are detected for each line in the ground truth (see image $\#$7 and $\#$9 in Table \ref{tab:result_images_part2}), whereas for other images, single lines are detected (e.g., image \#6 and \#8 in Table \ref{tab:result_images_part2}).

Among all methods, visually, our TGGLines performs the best. Another advantage of TGGLines is seen in Table \ref{tab:result_images_part2} image \#10 where multiple crossed lines form a circle intersection. As most methods detect lines based on edge maps, the circles from the original images leave an open circle shape in the results. While TGGLines avoids the open circle, and invariant to line width because we detect the lines based on image skeletons.

\subsection{Quantitative comparison}\label{sec:evaluation_quantitative}

We quantitatively compare the line segments from our TGGLines method and other four methods we benchmark against ground truth. Automatically quantifying the performance of detected line segment results is difficult, because the errors among the methods vary in nature. For example, some methods (like PPHT and LSD) detect most lines as double lines, but sometimes as single lines.

We use line detection accuracy as a simple evaluation metric, by manually counting the true positive line segments detected to calculate the accuracy. True positives are defined using the following criteria (examples of criteria are visually illustrated in Appendix \ref{app:evaluation_criteria_visual}): (1) for double line cases (e.g., PPHT, LSD), we count a line segment as correct if (a) both line segments are correctly detected compared with the corresponding line segment in the ground truth, so we assign weight 0.5 for each line segment; or (b) if more than half is detected for one segment, we give it weight 0.5 * 0.5 = 0.25; (2) for single line cases (e.g., TGGLines), we count a line segment as correct if (a) it is fully detected correctly, we assign weight 1; or (b) if more than half is correctly detected, we give weight 0.5; (3) if many tiny line segments are detected for a line segment in the ground truth, we view it incorrect, and assign it weight 0.

\begin{figure}[!htb]
\begin{center}
   \includegraphics[width=0.97\linewidth]{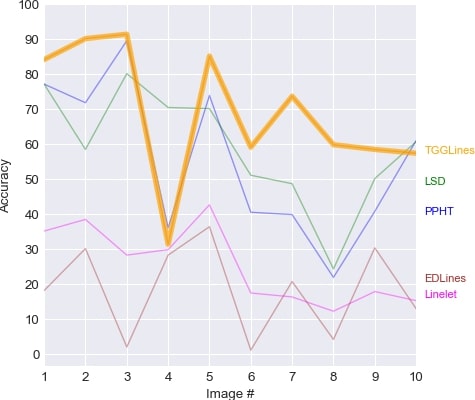}
\end{center}
   \caption{Line segment detection accuracy for different methods.}
\label{fig:results_comparison_accuracy}
\end{figure}

The accuracy calculation is based on the manually annotated line segments in ground truth. Specifically, \(accuracy = n_{c} / n_{t}\), where $n_{c}$ is the number of correctly detected line segments (true positive), and $n_{t}$ is the total number of line segments in ground truth.  For methods that (e.g., LSD and EDLines) detect double lines for each line in the ground truth, we only count it detected correctly if it detect both line segments. The accuracy of the methods on line segment detection is provided in Figure \ref{fig:results_comparison_accuracy}. We can see that the accuracy results match up with the visual comparison. EDLines performs the worst among the images in our dataset, and TGGLines performs the best, while Linelet is not consistent. 

While manually counting line detection results, we note that TGGLines is the only method that avoids detecting double lines, and it detects lines with the least break points.

\section{Conclusion}\label{sec:conclusion}


We have introduced a robust topological graph guided line segment detection method, \textit{TGGLines}, for low quality binary diagram images, using a skeleton graph image representation. Compared with state-of-the-art line segment detection methods (specifically, PPHT, LSD, EDLines, and Linelet), on diagram images taken from a patent image database; empirical results show that our TGGLines approach outperforms other methods, both visually and quantitatively.

Beyond the accuracy of TGGLines compared with other methods, TGGLines has two competitive advantages: (1) it is robust, as TGGLines only requires one parameter and most importantly, this parameter is \textit{adaptive}; and (2) the line segments detected using TGGLines are organized in a topological graph -- we name it \textit{line segment connectivity graph (LSCG)}. Thus the topological relations among the line segments are captured and stored while detecting line segments which can be used for further topological-based image analysis.

The effectiveness and robustness of our method, especially the topological relations among detected line segments, provide an important foundation for many applications, such as text recognition in historical documents and OCR-based applications, converting rasterized line drawings to vectorized images, road lane line detection for autonomous driving, as in real world road scene images often contain similar zigzag ``noise" as scanned document images have, because of scenarios such as worn road markings, falling leaves and/or dirt over roads.


{\small
\bibliographystyle{ieee_fullname}
\bibliography{__references_CVPR2020_lineDetect}
}


\newpage
\clearpage
\pagenumbering{arabic} 
\renewcommand*{\thepage}{A - \arabic{page}}

\appendix
\section{Appendix}




\subsection{Abbreviations}\label{app:abbr}


In this appendix, we provide the abbreviations (ordered alphabetically) of terms we used in the paper.

\noindent 
\begin{tabular}{@{}ll}

EDLines & Edge drawing lines\\
HT & Hough transform\\
LSCG & Line segment connectivity graph \\
LSD &  Line segment detector \\
OCR & Optical character recognition \\
PPHT & Progressive probabilistic hough transform \\
TGGLines & Topological graph guided lines

\end{tabular}


\subsection{Definition of terms used}\label{app:def_graph_terms}

In this appendix, we provide the definitions to some concepts (ordered alphabetically; referenced \cite{Yang_2019_CVPR_Workshops,yang2015generation, mlg2019_7,harris2008combinatorics,needham2019graph}) in graph theory and computational geometry we used in our TGGLines.

\begin{description}[style=nextline,font=\normalfont]
    \item[Convex hull] The convex hull of a finite set of points $S$ is the intersection of all half-spaces that contain $S$ . A half space in two dimensions is the set of points on or to one side of a line. Computing the convex hull of $S$ is one of the fundamental problems of computational geometry. 
    
    \item[Embedded graph] An \textit{embedded graph} is a graph that each node has (planar) coordinates so the graph can be drawn on a plane uniquely without any edge intersection or crossing. 
    
    \item [Graph] A \textit{graph} consists of a collection of nodes (also called \textit{vertices} or \textit{points}) and a collection of edges that connect the nodes.
    
    \item [Skeleton] The \textit{skeleton} of a binary image is a central line extraction (1 pixel wide) of objects present in the image through thinning.
    
    \item [Skeleton graph] The \textit{skeleton graph} $G_s$ of a binary image $B$ is an embedded graph generated from the skeleton $S$ of $B$, where each node in $G_s$ represents a white (i.e., non-zero) pixel in $S$ and each edge $e$ in $G_s$ indicates the two pixels represented by the end nodes of $e$ are neighbors.

    \item [Path] In graph theory, a \textit{path} is a sequence of distinctive nodes connected by edges. The length of a path is the number of edges traversed. node $v$ is reachable from node $u$ if there is a path from $u$ to $v$. A graph is connected, if there is a path between any two nodes.

\end{description}


\subsection{Parameter settings and computational time} \label{app:parameter_setting_computational_time}
This appendix provides the parameter settings for each method in Table \ref{tab:result_images_part1} and \ref{tab:result_images_part2}.

Table \ref{tab:parameter_settings} provides the parameters used for each method we compare. Computing environment for the experiments is provided in Appendix \ref{app:computing_env} below.




\begin{table*}
\caption{Parameter settings for different line segment detection methods in Table \ref{tab:result_images_part1} and Table \ref{tab:result_images_part2}. For other parameters not listed specifically here, default parameters provided in the tools are used.} \label{tab:parameter_settings}
\begin{tabular}{| L{3cm} | L{6cm} | L{5cm} |}

\hline
\textbf{Line segment detection methods} & \textbf{Parameters} & \textbf{Tools and methods used} \\
\hline

PPHT & \textit{threshold:} 10; \textit{line\_length:} 5; \textit{line\_gap:} 3   &  \textit{probabilistic\_hough\_line} function (Scikit-image) \\
\hline

LSD  & \textit{\_scale:} 0.8; \textit{\_sigma\_scale:} 0.6, \textit{\_quant:}  2.0, \textit{\_ang\_th:} 22.5, \textit{\_density\_th:} 0.7 & \textit{createLineSegmentDetector} function (OpenCV) \\
\hline

EDLines & \textbf{Internal parameters: }\{\textit{ratio:} 50, \textit{angle\_turn:} 67.5*np.pi/180, \textit{step:} 3\};\textbf{ Parameters for Edge Drawing:} \{\textit{ksize:} 3, \textit{sigma:} 1, \textit{gradientThreshold:} 25, \textit{anchorThreshold:} 10, \textit{scanIntervals:} 4 \}; \textbf{Parameters for EDLine}: \{\textit{minLineLen:} 40, \textit{lineFitErrThreshold:} 1.0\}
 &  EDLines Python implementation (\href{https://github.com/shaojunluo/EDLinePython}{Github code})\\
\hline

Linelet  & \textit{param.thres\_angle\_diff:} pi/8;  \textit{param.thres\_log\_eps:} 0.0;  \textit{param.est\_aggregation:} Kurtosis &  Matlab code by the linelet authors \cite{cho2017novel} \\
\hline

\textbf{TGGLines} & TGGLines requires \textbf{only one} parameter, and it is \textbf{adaptive} (see line 7 in Algorithm \ref{alg:simplification}). & Implemented by the authors of the paper\\
\hline

\end{tabular}
\end{table*}

\subsection{Computing environment} \label{app:computing_env}
In this appendix, we provide the computing environment that we ran our experiments. We ran all the experiments on a Windows desktop machine (Windows 10) with Intel(R) Core(TM) i7-8700 CPU @ 3.20GHz (6 cores and 12 logical processors; 32.0 GB RAM).


\subsection{Visually illustrated criteria for quantitative evaluation} \label{app:evaluation_criteria_visual}

In this appendix, we provide the criteria for judging the correctness of line segments (Table \ref{tab:eval_criteria}) along with a visual example showing the criteria applied to detected line segments for each method that we compare during our quantitative evaluation (Section \ref{sec:evaluation_quantitative}).

Using the criteria provided in Table \ref{tab:eval_criteria}, we judge and calculate how many lines are correctly detected against the ground truth. Then the accuracy is calculated. See Figure \ref{fig:TGGLines_workflow_illustrativeexample} for an example of the criteria used to judge the lines detected for each method and how the accuracy is calculated for image \#01 from Table \ref{tab:result_images_part1}.

\begin{table*}
\caption{Judging criteria for evaluation. GT refers to ground truth. The line segments in the example in Figure \ref{fig:TGGLines_workflow_illustrativeexample} are color-coded by weights as shown here. For double-line cases (see Figure \ref{fig:TGGLines_workflow_illustrativeexample} (c) for an example), we only count the segment correct if both lines are correctly detected against the GT. Thus, our scoring process is as follows: each one of the double lines, if they are correctly detected, the weight for each line is assigned 0.5 and thus the score for the ``whole line" in GT is 0.5 * 2 = 1.0. Likewise, if only half of one line in a double line case is detected correctly, the score is 0.5 * 0.5 = 0.25.}
\label{tab:eval_criteria}
\begin{tabular}{| L{9.5cm} | C{5cm} |}

\hline
\textbf{Criteria} & \textbf{Weights (marked with color)} \\
\hline

Fully correct (same slope as GT) & \hlc[orange]{1.0}  \\
\hline

A long line in GT detected as multiple split lines in results with all topological relationships well-preserved & \hlc[orange]{1.0} \\
\hline

Not fully correct but $>=$ 50\% is correct
 & \hlc[Magenta]{0.5} \\
\hline

Different slope as GT but most of the line(s) are correct
 & \hlc[Magenta]{0.5} \\
\hline

One line of \textit{a double line case} is correct & \hlc[Magenta]{0.5} \\
\hline

\textit{Multiple tiny line segments (including connected or not connected):} shifted a little bit & \hlc[Magenta]{0.5} \\
\hline

Only half of one line of \textit{a double line case} is correct & \hlc[green]{0.25} \\
\hline

One line of \textit{a double line case} is correct, but shifted a little bit among the multiple tiny line segments & \hlc[green]{0.25} \\
\hline

\textit{Multiple tiny line segments (including connected or not connected):} shifted a lot & 0 \\
\hline

\end{tabular}
\end{table*}

\begin{figure*}[!htb]
\begin{center}
    \begin{subfigure}{0.42\textwidth}
      \includegraphics[width=\textwidth]{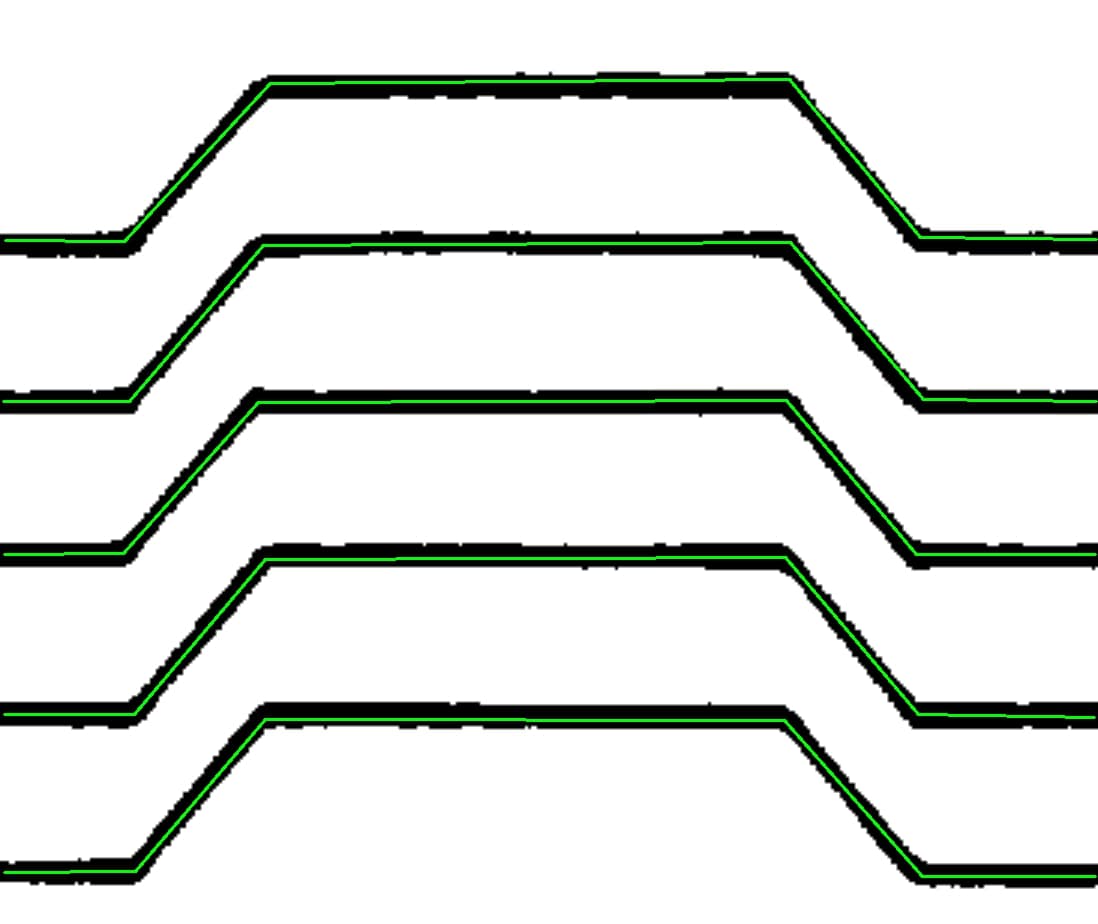}
      \caption{Ground truth: \# of lines = 25}
    \end{subfigure}
    \hfill
    \begin{subfigure}{0.42\textwidth}
      \includegraphics[width=\textwidth]{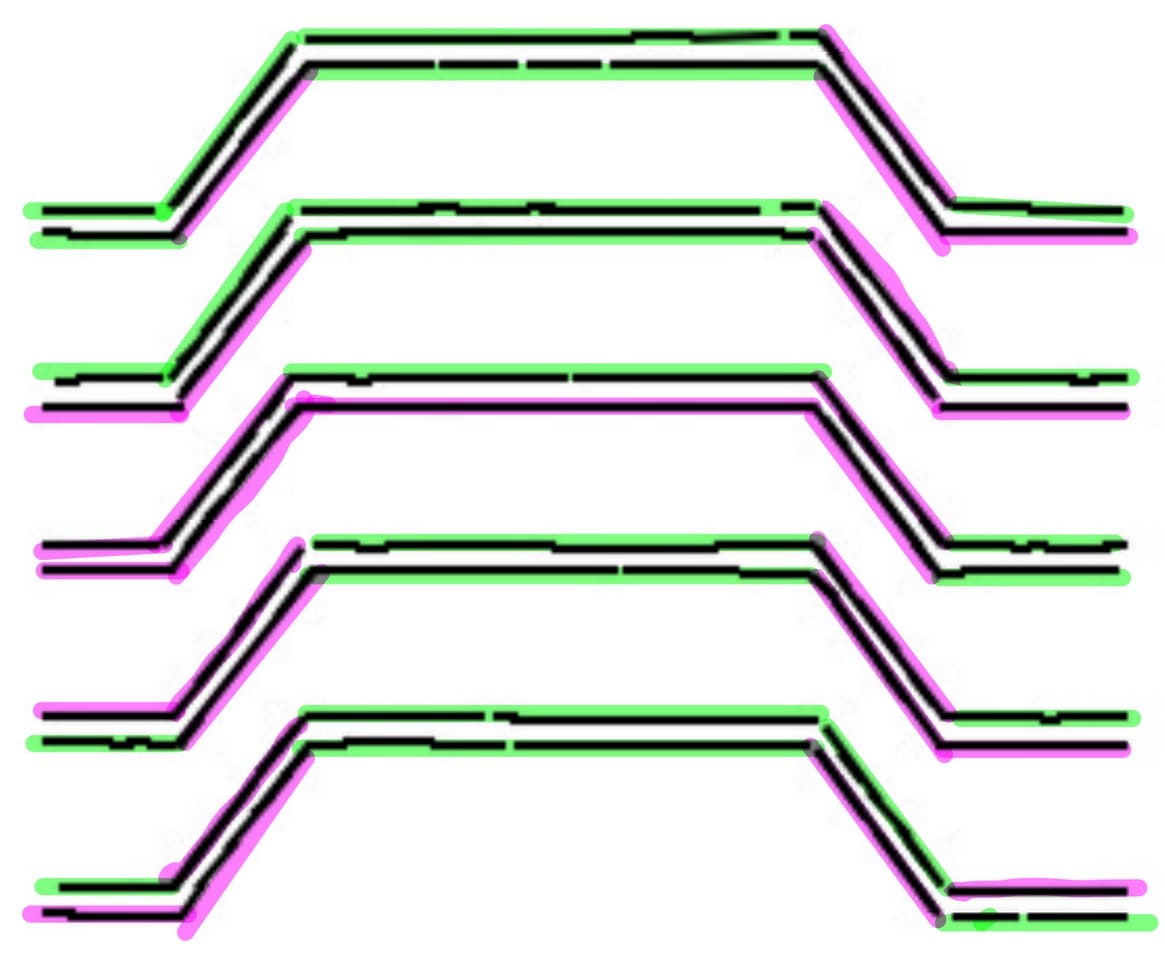}
      \caption{PPHT: \# of lines detected correctly = 27 * 0.5 + 23 * 0.25 = 19.25. Accuracy = 19.25 / 25 * 100\% =  77\%}
    \end{subfigure}
    \hfill
    \begin{subfigure}{0.42\textwidth}
      \includegraphics[width=\textwidth]{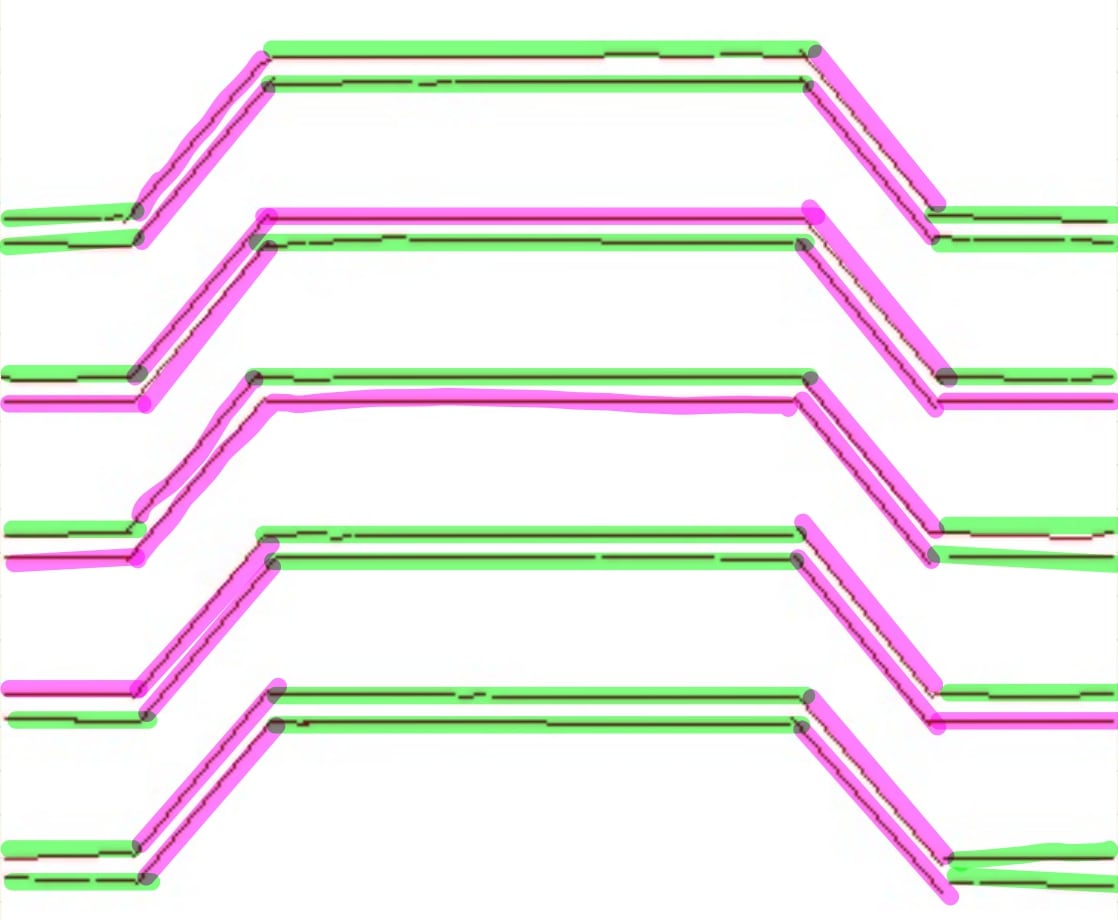}
      \caption{LSD: \# of lines detected correctly = 27 * 0.5 + 23 * 0.25 = 19.25. Accuracy = 19.25 / 25 * 100\% =  77\%}
    \end{subfigure}
    \hfill
    \begin{subfigure}{0.42\textwidth}
      \includegraphics[width=\textwidth]{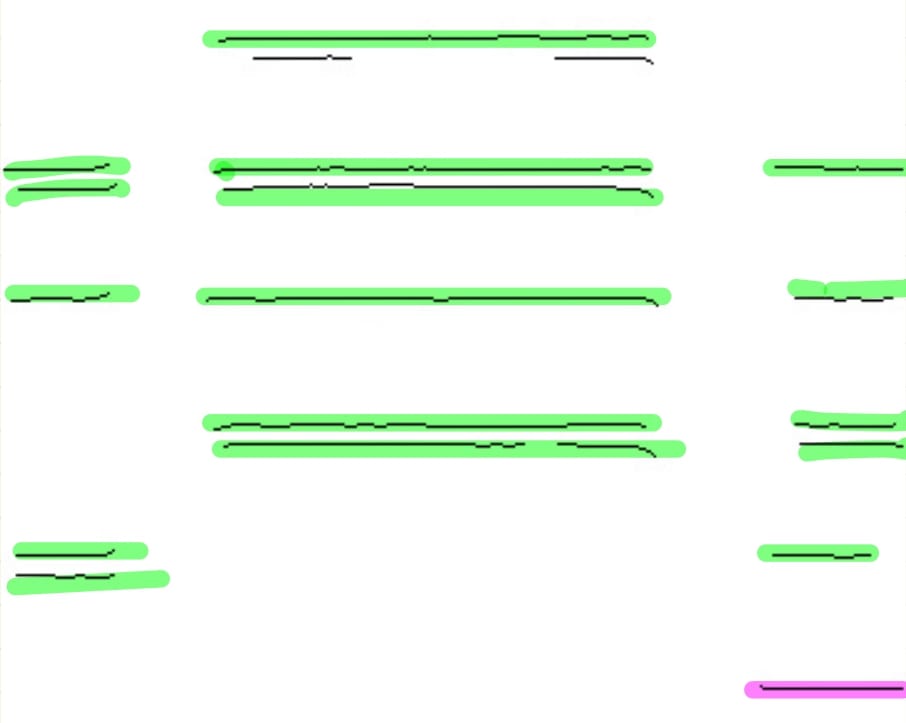}
      \caption{EDLines: \# of lines detected correctly = 1 * 0.5 + 16 * 0.25 = 4.5. Accuracy = 4.5 / 25 * 100\% =  18\%}
    \end{subfigure}
    \hfill
    \begin{subfigure}{0.42\textwidth}
      \includegraphics[width=\textwidth]{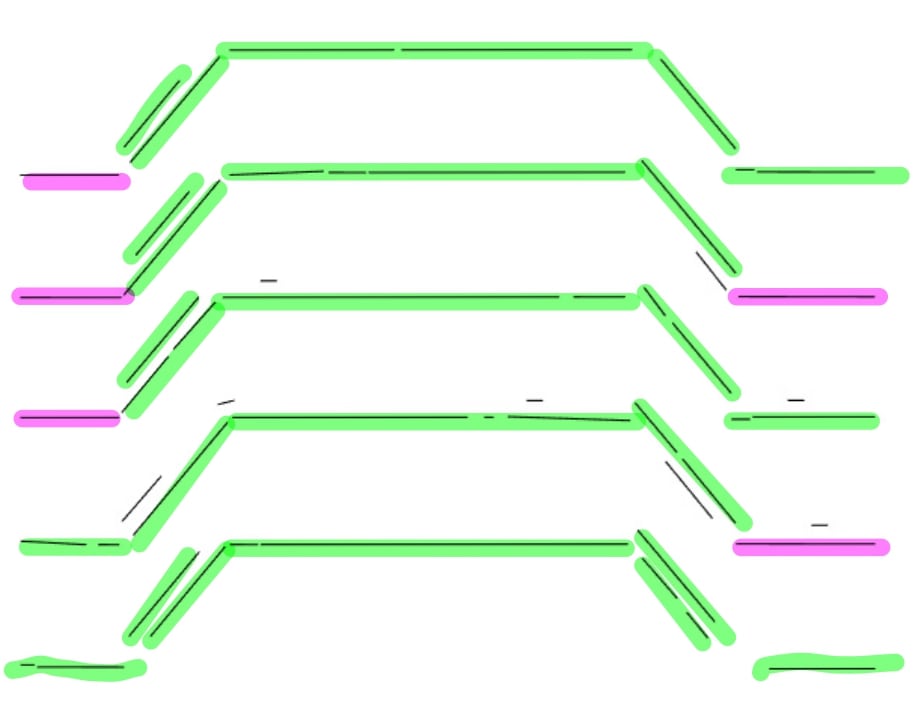}
      \caption{Linelet: \# of lines detected correctly = 5 * 0.5 + 25 * 0.25 = 8.75. Accuracy = 8.75 / 25 * 100\% =  35\%}
    \end{subfigure}
     \hfill
    \begin{subfigure}{0.42\textwidth}
      \includegraphics[width=\textwidth]{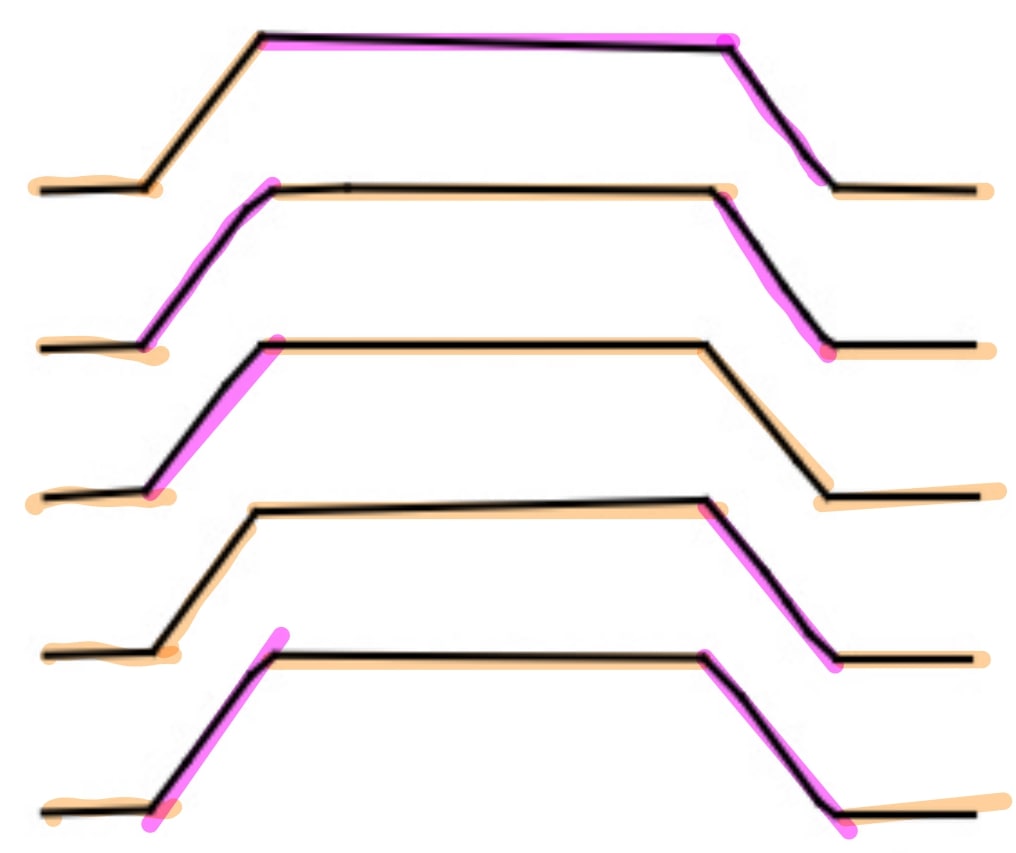}
      \caption{TGGLines: \# of lines detected correctly = 17 * 1 + 8 * 0.5 = 21. Accuracy = 21 / 25 * 100\% =  84\%}
    \end{subfigure}
\end{center}
   \caption{Judging criteria illustrated visually for the evaluation of different line segment detection methods for image \#01 in Table \ref{tab:result_images_part1}.} 
\label{fig:TGGLines_workflow_illustrativeexample}
\end{figure*}

\end{document}